\documentclass[12pt,a4paper]{article}

\PassOptionsToPackage{numbers, compress}{natbib}
\usepackage[preprint]{neurips_2023}

\usepackage[utf8]{inputenc} %
\usepackage[T1]{fontenc}    %
\usepackage{hyperref}       %
\usepackage{url}            %
\usepackage{booktabs}       %
\usepackage{amsfonts}       %
\usepackage{nicefrac}       %
\usepackage{microtype}      %
\usepackage{xcolor}         %

\usepackage{natbib}

\usepackage{hyperref}       %
\usepackage{booktabs}       %
\usepackage{amsfonts}       %
\usepackage{nicefrac}       %
\usepackage{microtype}      %
\usepackage{enumitem}
\usepackage{amsmath}
\usepackage{tabularx}       %
\usepackage{graphicx}
\usepackage{caption}
\usepackage{subcaption}
\usepackage{tcolorbox}
\usepackage{wrapfig}
\usepackage{multirow,booktabs,setspace,caption}
\usepackage{sidecap} %
\usepackage{floatrow}

\def\extractfont{\fontsize{8}{12}\selectfont\mathversion{normal}}

\newenvironment{extract}
{\list{}{}\item[]\extractfont}
{\endlist}

\graphicspath{{./figures/}}

\author{%
	Felix A.\ Wichmann\\
	University of Tübingen\\
	\texttt{felix.wichmann@uni-tuebingen.de} \\
	\And
	Robert Geirhos\\
	Google Research, Brain Team\\
	\texttt{lastname@google.com}
}

\begin{document}

	\title{Are Deep Neural Networks Adequate Behavioural Models of Human Visual Perception?}
	
	\maketitle
	
	\begin{abstract}
		Deep neural networks (DNNs) are machine learning algorithms that have revolutionised computer vision due to their remarkable successes in tasks like object classification and segmentation. The success of DNNs as computer vision algorithms has led to the suggestion that DNNs may also be good models of human visual perception. We here review evidence regarding current DNNs as adequate behavioural models of human core object recognition. To this end, we argue that it is important to distinguish between statistical tools and computational models, and to understand model quality as a multidimensional concept where clarity about modelling goals is key. Reviewing a large number of psychophysical and computational explorations of core object recognition performance in humans and DNNs, we argue that DNNs are highly valuable scientific tools but that as of today DNNs should only be regarded as promising---but not yet adequate---computational models of human core object recognition behaviour. On the way we dispel a number of myths surrounding DNNs in vision science.
	\end{abstract}

	\section{INTRODUCTION}
	Computational modelling in vision science has a long and rich history dating back at least to Schade's photoelectric analog of the visual system and Reichardt's motion detector \citep{Schade_1956,Hassenstein_1956,Reichardt_1957}. Computational models are useful because, first, a thorough and stringent quantitative understanding of an aspect of human visual perception implies we should be able to build a computational model of it. Second, a computational model can serve as a concrete, testable hypothesis of a theory and deepen our understanding through an iterative process of experimentation, model assessment and model improvement.
	Models in vision science come in many flavours, some borrowing computational elements such as filters or gain-control from engineering, some using information theory to motivate or derive properties of parts of the model, and others using Bayesian statistics to optimally read-out the activity within a model to derive the model decision: There is substantial diversity of models in vision science, and those models typically use whichever method or algorithm promises to be most helpful in a given context \citep{Cichy_2019}.
	
	In computer vision a new class of algorithms from machine learning (ML), so-called deep neural networks or DNNs, have effectively revolutionised the field. DNNs were explicitly developed to solve complex, real-world pattern recognition problems and are now often able to correctly identify an object in a ``typical'' real-world photograph. DNNs entered centre stage in 2012 when a DNN named AlexNet, designed and trained by \citet{krizhevsky2012imagenet}, comprehensively won the ImageNet Large Scale Visual Recognition Challenge (ILSVRC): Trained on 1.2 million images it was able to classify a set of 50,000 hitherto unseen images into 1,000 classes with an error rate of 16.4\%---down from 25.8\% in the previous year (by a non-DNN algorithm; error rates indicate top-5 error, i.e.\ whether the ground truth label was contained in the top five predictions of the model). Ever since 2012 the ILSVRC was won each year by a DNN, and DNNs are the de facto standard for pattern recognition in machine learning and computer vision\footnote{Since 2017 ILSVRC is not run anymore given that DNNs have basically solved the challenge: State-of-the-art (SOTA) top-5 performance is less than 1.0\% error. However, the ImageNet dataset still serves as a benchmark and is widely used as such.}
	In essence, a DNN is a nonlinear function approximator implemented as a very large collection of simple units which are connected to (typically a subset) of other units by variable connection strength, the so-called weights. The function of the DNN arises from the collective action of units, their connectivity pattern (the DNN architecture), the weights connecting units and their nonlinear activation function, i.e.\ how inputs to a unit are combined and transformed into its output passed to other units. DNNs are sometimes said to be based on neuroscience with units (neurons) connected to each other via variable weights (axons and dendrites with synapses of varying strength). The initial inspiration for neural networks did indeed come from neuroscience but DNNs only embody a highly simplified version of real neurons, their rich dynamics, intricate connectivity and dendritic processing complexity. The ``neural network'' part in the name of DNNs should thus be taken with more than a grain of salt \citep[c.f.][]{Douglas_1991}.
	Accessible introductions to DNNs are, for example, \citet{lecun2015deep,kriegeskorte2015deep} or \citet{serre2019deep}. A comprehensive treatment of DNNs can be found in \cite{goodfellow2016deep}.
	In a simplified cartoon-history, DNNs (with four or more layers) could be regarded as the children of (three-layer) connectionism popular from the mid 1980s to the late 1990s \citep{Rumelhart-etal_1986,McClelland-etal_1986}, and as the grand-children of the (two-layer) perceptron popular from the late 1950s to the late 1960s \citep{McCulloch-Pitts_1943,Rosenblatt_1958}\footnote{As we state, a highly simplified cartoon-history; we are aware of some of the other DNN milestones like the Neocognitron \citep{Fukushima_1980}, LeNet \citep{LeCun-etal_1989} or HMax \citep{riesenhuberHierarchicalModelsObject1999} which cannot be placed conveniently in the ``family history'' as presented above.}. A comprehensive overview of the more complex and multistranded historical development of deep learning is provided by \citet{Schmidhuber_2015}.

	\section{DEEP NEURAL NETWORKS AS TOOLS IN SCIENCE}
	\label{tools}
	DNNs are able to perform highly non-linear mappings from potentially very high dimensional inputs---like images with tens of thousands of pixels---to potentially high dimensional outputs---like the one thousand categories of ImageNet\footnote{In theory shallow, three-layer networks are already universal function approximators \citep{hornik1989multilayer,Hornik_1991}. In practice, however, with finite number of hidden units, finite datasets and computing power, shallow networks did not succeed in solving many interesting, large-scale or real-world problems. DNNs, on the other hand, are phenomenally successful at discovering such non-linear and high-dimensional mappings in practice.}. DNNs learn the non-linear and high-dimensional input-output mapping from massive amounts of training data alone: they learn which features and dimensions of the input, and their non-linear transformations and combinations, are relevant to solve a (classification) task. This ability to find non-linear, high-dimensional mappings makes DNNs a powerful tool in the sciences in general. Here we would like to draw a distinction between \emph{tools} and \emph{models}---knowing that the distinction is somewhat blurry at its boundary. Tools are a means to an end, they play an important role in the scientific process but are not, in itself, of scientific interest. In general a model, on the other hand, is a concrete instance of a theory and is itself of scientific interest. Many statistical algorithms or methods and tests are tools: We use linear regression, or analysis of variance (ANOVA) for scientific inference but, typically, we do not take them to be scientific models of the phenomenon under investigation.
	
	\definecolor{box.colbacktitle}{RGB}{190, 210, 165}
	\definecolor{box.colback}{RGB}{229, 237, 219}
	\begin{figure}[ht]
		\small
		\begin{tcolorbox}[width=\linewidth,colback={box.colback},
			title={\textbf{MODEL VS.\ TOOL}},
			colbacktitle=box.colbacktitle,coltitle=black]
			\textbf{Tool:} Algorithm or statistical method which plays an important role
			in the scientific process but is not, in itself, of scientific interest; a means to an end.\\
			
			\textbf{Model:} A concrete instance of a theory and itself of scientific interest. Different types of models exist such as statistical or mechanistic models as well as different modelling goals; see section~\ref{model}.
		\end{tcolorbox}
	\end{figure}
	
	As tools, DNNs excel in science: For example, DNNs have helped to speed up and improve single-molecule localisation microscopy \citep{Speiser-etal_2021}, make fast and accurate 3D protein folding predictions \citep{Senior-etal_2020}, substantially accelerate climate science simulations \citep{Ramadhan-etal_2022}, help in rapid gravitational wave parameter estimation \citep{Dax-etal_2021} and in applied mathematics deep reinforcement learning was used to find faster matrix multiplication algorithms \citep{Fawzi-etal_2022}.
	In the neurosciences and vision science DNNs as tools have led to clear improvements of methods: DeepLabCut allows markerless pose estimation of single \citep{Mathis-etal_2018} and multiple animals \citep{Lauer-etal_2022}, helping video-based observation and analysis of freely behaving animals. Simulation-based inference allows parameter inference in computational models that, without deep learning, would remain computationally intractable and has begun to see successful cognitive neuroscience applications \citep{Goncalves-etal_2020,Boelts-etal_2022}. \cite{Goetschalckx-etal_2021} argue that so-called generative adversarial networks (GANs) can be used to generate visual stimuli which are, on the one hand, complex and realistic but, at the same time, offer much more control than simply trying to find suitable stock or internet images.
	
	The success of DNNs as tools goes beyond method improvements, however. An important additional role is to use DNNs for exploration, as was clearly and convincingly argued for by \cite{Cichy_2019}\footnote{ We should note, however, that we differ from \cite{Cichy_2019} in terminology but not in substance: We refer to DNNs for exploration as tools, rather than models. Clustering algorithms or principal-component-analysis (PCA), for example, are also exploratory statistical tools. We prefer to term them tools rather than models because they are typically used as a means to explore rather than a model of the underlying phenomenon.}. One aspect of exploration are proof-of-concept or proof-of-principle demonstrations. \cite{Piloto-etal_2022}, for example, showed that important aspects of intuitive physics can be acquired entirely through visual, bottom-up, learning. In a similar vein in the domain of gloss perception, \cite{storrs2021unsupervised} used unsupervised DNNs to demonstrate how perceptual dimensions like gloss---only imperfectly corresponding to distal physical properties---can be learned from the entangled proximal stimulus without the need for (Bayesian) prior knowledge or generative models. DNNs can thus be used to explore how richly structured our visual environment is, and how much about the (distal) 3D world could in principle be extracted from (proximal) 2D sensors in a purely discriminative fashion \citep[see e.g.][]{Storrs_2021b}\footnote{
		One may speculate how much Gibson would have appreciated DNNs as proof-of-principle tools, as he argued throughout his career that the visual input alone---the optic array---is sufficiently rich to allow visual behaviour \citep{Gibson_1950a} whereas many Bayesian or predictive coding accounts of vision typically stress the---allegedly---impoverished nature of the visual input requiring prior knowledge and/or generative processes to accomplish visual perception. DNNs have clearly demonstrated that much more visual information can be obtained from images or videos alone than some vision scientists may have presumed.}. Another exploratory aspect of DNNs is the generation of new hypotheses: In beautiful work \cite{Rideaux-etal_2021} used a neural network to find a causal role for neurones in the macaque medial superior temporal (MST) area whose tuning properties had hitherto been regarded as puzzling. Their neural network analysis suggests that those neurones may play a role in the decision whether self- and scene motion signals arise from the same source or not---and should thus be either combined or analysed separately by downstream neurones.  
	The usefulness of DNNs as tools in (vision) science is beyond doubt; their immense predictive power in high-dimensional input-output mappings is crucial for advanced method acceleration and development, for stimulus generation and for exploration as in proof-of-principle demonstrations or the generation and test of new computational hypotheses.
	
	Sometimes it is tempting to turn successful statistical and computational tools into theories---the \emph{tools to theory heuristic} \citep{Gigerenzer_1991}. Scientists use the tools they use when working as metaphors for the workings of the phenomenon under study. Not perhaps in case of linear regression or ANOVA mentioned above, but a more general example mentioned by Gigerenzer is, e.g., the idea of the mind as a statistician by Brunswik, after Pearson had developed inferential statistics. In vision science a modern successor to this idea is the well-known Bayesian brain hypothesis: Bayesian statistics are a useful tool to analyse experimental data and combine them with prior beliefs. The Bayesian brain hypothesis asserts that the visual system itself applies the Bayesian probability calculus to sensory inputs---not only the scientist trying to understand the brain \citep[c.f.][]{Zednik_2016}. Given their immense usefulness as tools in general, and their success in object recognition in computer vision in particular, it is thus perhaps not surprising that DNNs have also been proposed as computational models of human (core) object recognition in vision science \citep[e.g.][]{Yamins2014,kriegeskorte2015deep,kubilius2019brain}\footnote{In case of DNNs it may even be a \emph{theory-to-tools-to-theory} heuristic as the development of neural networks was initially inspired by the visual brain, and now DNNs are re-imported into vision science---Frank Jäkel, personal communication, 02.11.2022.}.
	We will evaluate DNNs as models of human core object recognition in section~\ref{assessmentDNN}; first we will have a look at what we require of a good model in the next section.
	
	\section{WHAT MAKES A GOOD MODEL A GOOD MODEL?}
	\label{model}
	As stated in the introduction, computational models are useful because, first, they force scientists to make assumptions explicit and specify dependencies within a model more precisely than is possible by language alone \citep[c.f.][]{brickIllusoryEssencesBias2021}. Second, based on prior knowledge and previous insights they serve as concrete, experimentally testable hypotheses of theories and should explain data, processes or how processes interact.
	To model visual behaviour we traditionally employ mechanistic models: Models that capture human behaviour in terms of inputs and outputs, and whose computational ingredients are domain specific, i.e.\ informed by, or derived from, basic principles known and established in vision science. Typical mechanistic models of spatial vision, for example, employ spatial filters and divisive contrast gain-control as some of their computational and causal building blocks \citep[e.g.][]{Goris2013,Schutt_2017}. Mechanistic models in vision science come in different flavours, depending on how much they focus on behaviour versus how much they focus on neurophysiological realism. Mechanistic models of psychophysics are abstracted away from the details of the biological implementation---without, hopefully, being outright neurally implausible. If neural plausibility or realism is one of the goals of modelling, the models' ingredients or computational building blocks closely mimic the neuronal hardware and are linked to behaviour by linking propositions \citep{Teller_1984}.
	Statistical models, on the other hand, are only concerned with fitting and predicting data, that is, with the correct mapping between inputs and outputs. Typically they are generic and can be used in many different domains. DNNs used as tools as described in section~\ref{tools} are a prime example of statistical models. However, we still prefer to refer to them as tools if used as a means to an end only, in order to separate this use from the way they are also claimed to be (statistical) \emph{models} of core object recognition (the ventral stream). Importantly, in the absence of any concrete computational domain knowledge, statistical models are often the only models one could possibly use and they thus often precede mechanistic models.

	\definecolor{box.colbacktitle}{RGB}{190, 210, 165}
	\definecolor{box.colback}{RGB}{229, 237, 219}
	\begin{figure}[ht]
		\small
		\begin{tcolorbox}[width=\linewidth,colback={box.colback},
			title={\textbf{STATISTICAL VS.\ MECHANISTIC MODEL}},
			colbacktitle=box.colbacktitle,coltitle=black]
			\textbf{Statistical Model:} Concerned with fitting or predicting data, a correct input-output mapping; probabilistic model describing the relationship between the independent (input) and dependent (output) variables; general purpose.\\
			
			\textbf{Mechanistic Model:} Attempts to find a mechanistic, causal relationship between inputs and outputs which predicts the data; typically uses domain specific components, processes or transformations.
		\end{tcolorbox}
	\end{figure}

	Apart from different scientific goals leading to different types of models, there are multiple modelling desiderata---what we expect our model to be good at, whatever type of model it may be. For mechanistic models of behaviour the following desiderata are of particular importance: how well the model fits the data (``predictivity''), and how much the model aids understanding (``explanation''). In the following we expand on these modelling desiderata and discuss how well, in general, DNNs \emph{as models} fulfil them. 
	
	\subsection{Predictivity}
	\label{predictivity}
	Traditionally \emph{goodness-of-fit} was assessed when modelling: How well are my (already collected) data described by my model. Furthermore, if several models were compared, \emph{model selection} was applied to select the model offering the best trade-off between fitting the data and model complexity. With the advent of modern and highly adaptable ML classification algorithms the focus shifted from explaining past data (goodness-of-fit) to a more stringent test, namely predicting new data (often termed generalisation in ML). DNNs are trained---their parameters are optimised---on \emph{training} data but their performance is assessed how well they \emph{predict} the hitherto unknown \emph{test} data. On standard datasets such as ImageNet, or when used as tools in science, DNN prediction performance is often spectacularly good and the reason why DNNs are currently so popular. Predictivity can be measured at different levels: at an aggregate level (e.g.\ ``do models achieve human-level accuracy?''), but also at a more fine-grained level, for instance asking whether models predict human response and error patterns at an individual stimulus/image level, which is a much stricter requirement \citep[c.f.][]{green1964consistency}. In the case of classification data, the latter can be measured by \emph{error consistency} \citep{geirhos2020beyond}, an image-level metric to assess the degree of similarity between for instance human and machine error patterns: Whether humans and machines agree on which images are easy or difficult to classify \citep[also see][]{rajalingham2018large}.
	
	Good prediction performance, ideally both at the coarse and the fine-grained level, is undoubtedly an important desideratum of a good model. However, we argue below that it is only one aspect of a good model. Using prediction performance on a specific task as a \emph{benchmark} is exceedingly popular in computer vision and ML, and benchmarks are gaining traction in vision science, too. While benchmarks have fuelled a lot of progress, an over-reliance on benchmarks to measure progress can be problematic if model quality is erroneously thought of as a single dimension along which models can be ranked, turning science into a spectator sport; as we argue here, model assessment is multi-dimensional and a single rank does not do good models justice. Furthermore, benchmarks and rankings encourage small and short-term gains and discourage fundamental re-thinking and the exploration of novel directions which, typically, are initially accompanied by worse performance on a benchmark.

	\subsection{Explanation}
	Beyond predictivity, an important desideratum of a good model is that it helps to \emph{explain} the data and how the building blocks of it are causally linked to the observed behaviour. Thus, a good model helps explain a scientific phenomenon, or at least aids in its interpretation and understanding.
	One way models act as an explanation and aid understanding is if they are simple---they have a limited number of parameters and contain building blocks or modules with identifiable sub-functions. Many of the mechanistic models in vision science are of this type and they appear to follow the adage of George Box:
	\begin{extract}
		Since all models are wrong the scientist cannot obtain a ``correct'' one by excessive elaboration. On the contrary following William of Occam he should seek an economical description of natural phenomena. Just as the ability to devise simple but evocative models is the signature of the great scientist so overelaboration and overparameterization is often the mark of mediocrity. \cite[p.~792]{Box_1976}
	\end{extract}
	DNNs are made up from simple, well-understood, computational units, but their decisions result from the interplay of hundreds of thousands of units and millions of connections. As of today, DNN decisions remain largely opaque because the post-hoc methods developed to understand DNNs---and thus provide an explanation for their behaviour useful to a scientist---have not yet matured enough: Neither visualising maximally activating features \citep{Gale-etal_2020,Borowski-etal_2021,Zimmermann-etal_2021} nor heatmap or saliency methods \citep{Kindermans-etal_2017,Montavon-etal_2017,adebayo2018sanity} sufficiently explain the functions learned by DNNs. While the methods may still mature and there also exist promising novel approaches which may help in the future, e.g.\ \cite{Cohen-etal_2020} or \cite{Chung-Abbott_2021}, for now DNNs provide not as much of an explanation and do not aid understanding as much as we would like them to---they should thus at best be regarded as \emph{statistical} models in vision science. 
	It should not be left unmentioned, however, that it is not inconceivable that models of an organ as complex as the brain, and behaviours as complex as, for example, object recognition, cannot be modelled using a simple model---at least not whilst having adequate prediction performance. Highly nonlinear systems may require highly nonlinear models and both are notoriously difficult to understand and analyse. This is true for complex nonlinear mechanistic models as it is true for statistical models like DNNs. For successful models of visual perception there may not be a dichotomy between understandable ``traditional'' (mechanistic) models on the one hand and impenetrable (statistical) DNNs on the other: It is a matter of the complexity of the behaviour our model should predict.
	Thus however much we like simplicity it may well be that if we want computational models of complex visual behaviour we have to come to terms with Santiago Ramón Y Cajal's insight into nature:
	\begin{extract}
		Unfortunately, nature seems unaware of our intellectual need for convenience and unity, and very often takes delight in complication and diversity. \citep[p.~240]{Cajal_1906}
	\end{extract}
	Ideally, models should be highly predictive, show human-like error patterns and aid understanding. In practice, all those desiderata appear difficult to achieve when modelling complex human visual behaviour. Different models may be preferable depending on the modelling purpose: DNNs if a highly predictive model is desirable even if it does not provide much of an explanation. At other times a simple, easy to understand mechanistic model which is not nearly as predictive may still be preferable. Thus a model may not only be either good or bad but can be both: Depending on the purpose, models can simultaneously be great and poor, useful and futile: Model assessment is multi-dimensional.
	
	\section{ADEQUATELY MODELLING VISUAL CORE OBJECT RECOGNITION}
	\label{assessmentDNN}
	Model assessment is multi-dimensional---but how adequate are current DNNs as models of human perception? We here restrict ourselves to \emph{core} object recognition \citep{dicarlo2012does}: Our fast and seemingly effortless ability to classify objects in the real world or images (photographs) of objects to belong to a particular class or category at the entry level---a cat, elephant, car, house, aeroplane etc. Clearly, human object recognition is more than only classification (and thus more than core object recognition): We are also able to recognise individual exemplars from large and more or less homogenous classes, most obviously in face recognition \citep{OToole-Castillo_2021} but also when recognising our cat amongst a dozen other cats in the garden, our car in a parking lot or our bike in the overcrowded bike-rack in front of a University building \citep[for a review see, e.g.][]{Logothetis_1996}. Obviously, there is also more to human vision than object recognition: We use vision to estimate properties of materials and scenes, assessing, for example, distances and surface angles, or use it to guide behaviour and to navigate. But notwithstanding how much else there is to vision, (core) object recognition is undoubtedly very important for perception:
	\begin{extract}
		At a functional level, visual object recognition is at the centre of understanding how we think about what we see. Object identification is a primary end state of visual processing and a critical precursor to interacting with and reasoning about the world. \citep[p.\ 76]{Peissig_2007}
	\end{extract}
	Furthermore, when assessing the adequacy of DNNs as models of human perception it is fair and appropriate to use a task that is not only important for humans and computationally complex, but also one where DNNs excel at. In computer vision, object recognition is continuing to set the standards for DNN performance. The combination of these factors---the importance of object recognition for human perception, the central role of object recognition within computer vision, and the fact that DNNs trained on object recognition are being proposed as models for primate ventral stream core object recognition---collectively render visual core object recognition perhaps the best task for comparing human against machine behaviour and thus assessing the adequacy of DNNs as models of a particularly important aspect of human visual perception.
	What needs to be successfully modelled in core object recognition are the following central, often replicated and most important findings \citep{Biederman_1987}: The (rapid) recognition of objects under changes in orientation, illumination, partial occlusion as well as if distorted by (moderate levels of) visual noise.
	
	\definecolor{box.colbacktitle}{RGB}{156, 174, 211}
	\definecolor{box.colback}{RGB}{217, 222, 238}
	\begin{figure}[ht]
		\small
		\begin{tcolorbox}[width=\linewidth,colback={box.colback},
			title={\textbf{MYTH: ALEXNET IS A GOOD DEFAULT MODEL}},
			colbacktitle=box.colbacktitle,coltitle=black]
			In vision science, AlexNet \citep{krizhevsky2012imagenet} is still one of the most commonly investigated deep learning models: The number of abstracts containing the word ``AlexNet'' was 2.6 times higher than the number of times a ``transformer'' was mentioned at the 2022 conference on vision science, VSS (13 vs.\ 5). While there are historical reasons for this---after all, AlexNet was the very network that started the deep learning revolution back in 2012---we believe that more than a decade later there are hardly any contemporary reasons for using AlexNet as the default model in an investigation. From a vision science perspective, there are better models than AlexNet on most metrics (distortion robustness, texture/shape bias, error consistency, adversarial robustness). From a computer vision perspective, AlexNet has been outdated for nearly a decade now: AlexNet used to be a commonly employed model from 2012 till at the very latest 2014; ResNets \citep{he2015delving} were the leading model family from 2015 till 2020 and vision transformers \citep{dosovitskiy2020image} are now rapidly becoming the new standard since 2020. We believe that retiring AlexNet in favour of more modern DNN families (often trained on large-scale datasets) is likely to lead to more insights and discoveries.\\
			
			We suspect that AlexNet may still be popular in vision science because, first, it appears ``intuitively'' closer to the neurophysiological roots of DNNs: Unlike modern transformers, for example, AlexNet uses convolutions, unlike very deep DNNs AlexNet has ``only'' 8 layers---in the ballpark of the ventral stream---and it has two processing streams, perhaps like M- and P-pathways (or the dorsal and ventral streams?)---even if AlexNet only has two streams to allow for an efficient implementation on two GPUs. Second, by modern standards AlexNet is a ``small'' DNN perhaps giving rise to the hope it may be easier to analyse and may even serve as a \emph{mechanistic} model of human vision. We do not believe either argument to be convincing, however. First, AlexNet's similarity to the human brain is at best very superficial only, which becomes clear from its behavioural failures. Second, AlexNet may be a small DNN but it still has more than 650K units and 60M parameters---thus it remains an opaque network and a statistical model only (see section~\ref{model}).\\
			
			If AlexNet is to be included in an investigation it is advisable to compare it against state-of-the-art alternatives. The answer to whether DNNs adequately model human visual perception can only ever be a snapshot in time---but we should do our best in making sure it captures the present time, rather than the past.
		\end{tcolorbox}
	\end{figure}

	\subsection{Robustness to 3D viewpoints}
	\label{3Dviewpoints}
	\cite{Yamins2014} performed an impressive and influential set of experiments in which they optimised the \emph{architecture} of their DNNs with respect to the performance of the DNN in their object recognition task. One of their main findings---not central to our discussion here---was a correlation between performance of the DNNs on their object recognition tasks and the DNNs' ability to also predict neuronal firing patterns in monkey inferior temporal cortex (IT). The stimuli used by Yamins et al.\ to test their DNNs were 64 rendered objects (8 exemplars from each of 8 categories) superimposed on natural images as background (note that the objects and their position were not matched to the background, however, resulting in, for example, a boat or table floating in mid air above a landscape). What is central to the current discussion is that Yamins et al.\ varied the object views: From easy, canonical views in the low variation condition to strong changes of the orientation of the objects in the high variation condition. We know that one of the strengths of human object recognition is its ability to cope with changes in orientation, as mentioned above. Not surprisingly, thus, observers performed reasonably well also in the high variation condition \cite[Fig. 2B, p.\ 8621]{Yamins2014}. However, not only did human observers perform reasonably well in the high variation condition, but also their best DNN performed nearly as well (resulting from hierarchical modular optimization, HMO). This result suggested that DNNs may thus be on par with humans in their robustness to viewpoint or object orientation changes.
	
	\begin{figure}[h!]
		\includegraphics[width=\textwidth]{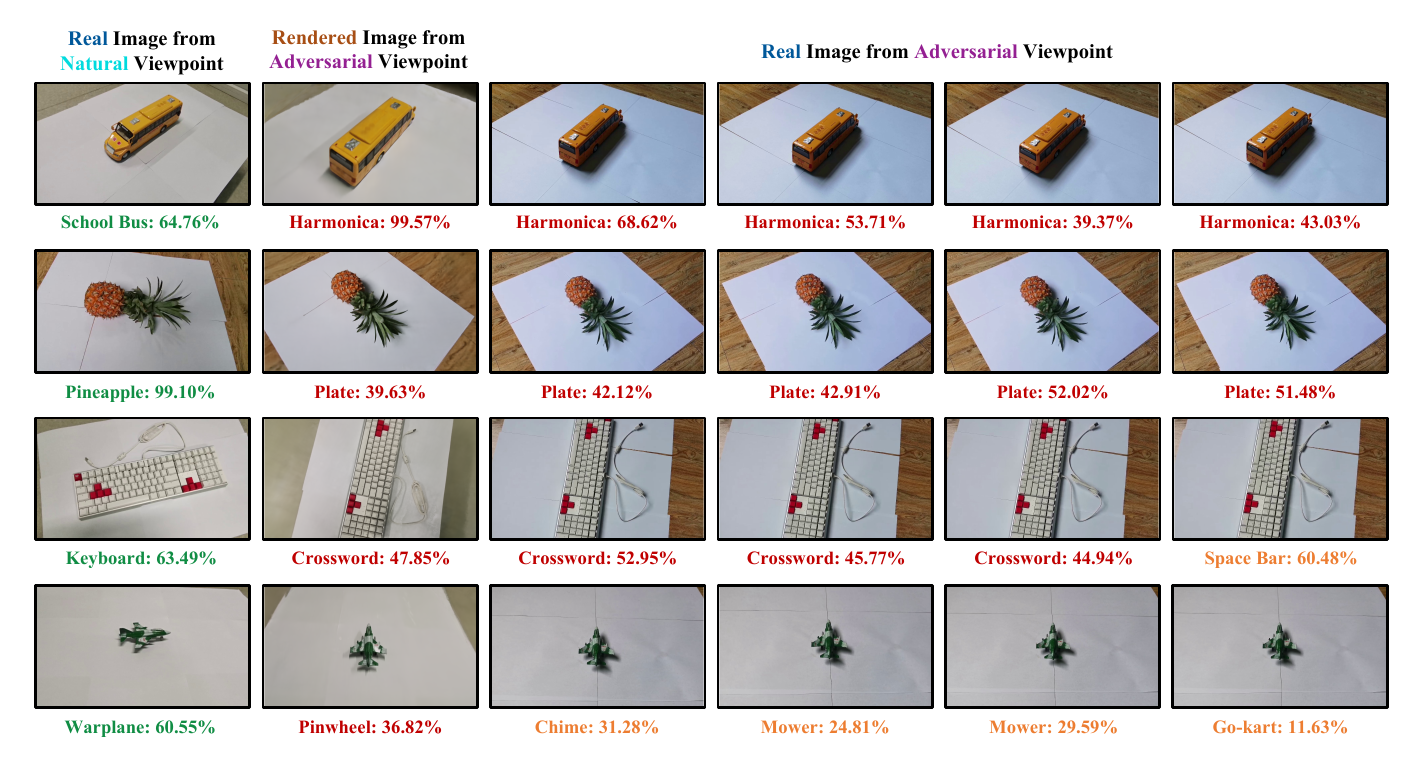}
		\hfill
		\caption{Visualisation of 3D viewpoint dependence of object categorisation of DNNs. The first column shows the
			real images from a ``natural'' viewpoints (correct classification; green label below images). The second column shows the rendered images from a viewpoint optimised to lead to wrong classification (``adversarial viewpoints''). Columns 3 to 6 show real images again but taken to approximate the adversarial viewpoints of column 2 (wrong classification; red or orange labels depending on DNN confidence in classification). \textit{Figure source: \cite{Dong-etal_2022}, Fig.\ 5, page 9 with kind permission from the authors.}}
		\label{fig:viewfool}
	\end{figure}
	
	However, the stimuli used by \cite{Yamins2014} were, first, comparatively few and, second, not embedded in a natural background but only superimposed. \cite{Dong-etal_2022} elegantly explored the viewpoint dependence of DNN object recognition systematically for a number of modern DNNs (VGG-16, Inception-v3, DenseNet-121 and five others). They found classification accuracy of all DNNs to be 3D view dependent---and strongly so: Whilst classification accuracy was typically between 70 and 80\% for standard ImageNet images, performance dropped dramatically with slightly unusual viewpoints to \emph{below} 20\% for most DNNs, with the best performing DNN still below 50\% accuracy\footnote{\cite{Dong-etal_2022} did not simply search for unusual viewpoints but optimised them to find views that are difficult for DNNs, a procedure similar to how adversarial images are generated, see section \ref{adversarial}. This is the reason the authors call them \emph{adversarial viewpoints}.}. Note that none of the problematic viewpoints for DNNs pose any difficulty for human observers: The difficult images for DNNs are quickly and effortlessly recognisable for us---at least the examples shown in the paper, see Fig.~\ref{fig:viewfool}.
	Similar results were obtained in related studies by \citet{alcorn2019strike}, \citet{abbas2022progress} and \citet{ibrahim2022robustness}. Alcorn et al.\ concluded that their work ``revealed how DNNs’ understanding of objects like `school bus' and `fire truck' is quite naive---they can correctly label only a small subset of the entire pose space for 3D objects'' (p. 8). It is as well to note that also the images used by all the studies cited above do not contain occlusions; the tested DNNs sometimes generalise poorly to novel 3D viewpoints despite seeing all the relevant objects in full view. In this regard, a 3D viewpoint robustness gap between humans and machines remains.
	
	\subsection{Robustness to image distortions}
	\label{imageDistortions}
	Core object recognition performance should not only be (largely) viewpoint invariant but also robust against other image distortions such as, for example, moderate levels of visual noise. \cite{geirhos2018generalisation} systematically explored this issue for three DNNs (ResNet-152, VGG-19 and GoogLeNet) using 13 different image distortions or degradations. To several of the image distortions DNNs were clearly much less robust than the psychophysically tested human observers; the discrepancy was particularly pronounced for uniform noise, low-pass and high-pass filtering and the so-called Eidolon distortions \citep{Koenderink_2017}. In the case of uniform noise, for example, human observers were still around 50--60\% correct at a noise variance for which all tested DNNs were essentially at chance performance (6.25\% in a 16-fold identification task). Including the image distortions in the training of the DNNs led to super-human performance for the DNNs on the very distortion included in the training, but there was little to no generalisation to other distortions: Even training on undistorted images and images with low contrast, low-pass and high-pass filtering, phase noise, image rotation and salt-and-pepper noise did not help the DNNs to cope with uniform noise. Thus \cite{geirhos2018generalisation} concluded there is a large robustness gap between DNNs and human observers in their core object recognition ability in the face of image distortions, a conclusion that is corroborated by many other robustness studies \citep[e.g.][]{berardino2017eigen,wichmann2017methods, hendrycks2019benchmarking, koh2021wilds,hendrycks2021many, huber2022developmental, Idrissi-etal_2022}.
	
	However, progress in deep learning is sometimes remarkably swift and thus \cite{geirhosPartialSuccessClosing2021} re-assessed robustness to image distortions---termed out-of-distribution (OOD) robustness in ML---in 52 ``classic'' and state-of-the-art (SOTA) DNNs of 2021 using a total of 17 OOD datasets and more than 85.000 psychophysical trials in the laboratory. Overall different DNN architectures and training regimes appear to have little systematic influence on DNN robustness, but DNNs trained on at least 14 million images showed near human OOD robustness, with two models even surpassing humans in an aggregate measure of OOD robustness across the 17 datasets. What remained, however, was that none of the DNNs comes close to human robustness to low-pass filtering, consistent with all what we know about the different image features used by DNNs and humans during object recognition and discussed in the next section~\ref{imageFeatures}. We should also add that most of the DNN architectures popular in vision science---for example AlexNet, VGG, ResNets, DenseNets, Inception---show poor and clearly sub-human OOD robustness if only trained on ImageNet. The lack of OOD robustness of these DNNs should be kept in mind if one compares one or some of them to human performance or neural systems.
	Furthermore, while increased OOD robustness through training on large-scale datasets is indeed an impressive achievement of ML, we require more than just similar overall prediction performance if we want to assess DNNs as adequate model of human behaviour: We also require \emph{error consistency} even for ``only'' statistical models, as we argued in section~\ref{predictivity} above. With respect to error consistency there remains a large gap between \emph{all} of the 52 DNNs and human behaviour: Whilst the 90 human observers showed large agreement which images they felt were easy or difficult to recognise, human-machine error consistency is only at about half the value of human-to-human consistency even for the best DNNs \cite[see Fig.~1 (d), p.~5, in][]{geirhosPartialSuccessClosing2021}. Given the large remaining gap in error consistency between SOTA DNNs and human observers it appears safe to conclude that even the highly OOD robust DNNs appear to process images differently from human observers.

	\subsection{Image features underlying object recognition}
	\label{imageFeatures}
	Initially it was widely believed that DNNs recognise objects based on their shape---similar to how humans recognise objects: ``[T]he network acquires complex knowledge about the kinds of shapes associated with each category. [...] High-level units appear to learn representations of shapes occurring in natural images'' \citep[p.\ 429]{kriegeskorte2015deep} or that intermediate DNN layers recognise ``parts of familiar objects, and subsequent layers [...] detect objects as combinations of these parts'' \citep[p.\ 436]{lecun2015deep}.
	However, systematic investigations now seriously challenge this view that DNNs, like humans, recognise objects via their shape. \cite{baker2018deep} conducted a series of experiments in which DNNs (AlexNet and VGG-19) had to classify silhouettes of objects, silhouettes filled with the texture of another object, glass figurines and outline figures. For all stimulus variations DNNs performed poorly, suggesting that DNNs rely much more on surface characteristics like texture than human observers and lack global shape sensitivity. Similar results were obtained by \cite{brendel2019approximating} who successfully trained a ResNet-variant they termed ``BagNet'' on ImageNet. BagNet exclusively relies on a bag-of-local features---no shape encoding possible---but still performed similar to several standard DNNs like VGG-16, ResNet-152 or DenseNet-169 in terms of interactions between image parts, sensitivity to features and errors.
	
	Whilst the previous two studies assessed only DNN performance---either to changed stimuli or using a DNN architecture only capable of using local image patches for classification---in our own work we compared human behaviour directly with DNN classifications (``DNN behaviour''). We used style-transfer algorithms \citep{Gatys2016,Huang-Belongie_2017} to create cue-conflict stimuli, combining the shape of one object with the local surface characteristics (texture) of another, e.g. the shape of a cat with the texture (or skin) of an elephant \citep{geirhos2019imagenettrained}. Human observers in a large-scale psychophysical study predominantly classified the cue-conflict stimuli almost exclusively according to their shape, whereas ImageNet-trained DNNs showed a strong texture bias (57.1\%--82.8\% depending on architecture), indicating that they do not classify objects according to their shape as humans do. Importantly, human-machine comparisons need to be careful and fair: To ensure a core object recognition comparison in this and all our behavioural experiments presentation time was limited to 200 ms (a single fixation) and all images were immediately followed by a high-contrast 1/f noise mask to minimise, as much as psychophysically possible, feedback influences on perception. Finally, observers had to respond quickly to ensure perceptual rather than cognitive responses: core object recognition.
	
	\definecolor{box.colbacktitle}{RGB}{190, 210, 165}
	\definecolor{box.colback}{RGB}{229, 237, 219}
	\begin{figure}[ht]
		\small
		\begin{tcolorbox}[width=\linewidth,colback={box.colback},
			title={\textbf{SHAPE VS.\ TEXTURE BIAS}},
			colbacktitle=box.colbacktitle,coltitle=black]
			\floatbox[\capbeside]{figure}[\FBwidth]{
				\caption*{\textbf{Shape bias:} The tendency to classify images according to their global shape (``cat'' in the example here).\\
					
					\textbf{Texture bias:} The tendency to classify images according to their local texture-like characteristics (``elephant'' in the example here).\\\textit{Image source: \cite{geirhos2019imagenettrained} with permission from the authors.}}}
			{\includegraphics[width=0.2\textwidth]{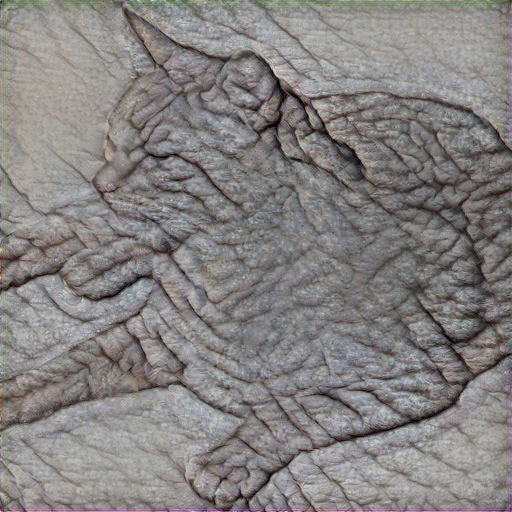}}
		\end{tcolorbox}
	\end{figure}

	The above result still held at least in 2021 when \cite{geirhosPartialSuccessClosing2021} re-assessed, amongst other performance measures, shape and texture bias of 52 DNNs. Amongst the 52 DNNs were a large variety of different SOTA DNN architectures and training variants, including adversarial training, selfsupervision and image-text training (CLIP). Some modern DNNs---adversarially trained, CLIP---exhibited a more human-like (stronger) shape bias than previous DNNs, but even they remain substantially more texture biased than human observers \citep[Fig.\ 3]{geirhosPartialSuccessClosing2021}. A recent study by \citet{Malhotra-etal_2022} again confirms that humans, but not the DNNs investigated, have a shape bias even when learning novel stimuli for which texture is more predictive. The authors argue that the human shape bias is likely resulting from an inductive bias for shape \citep[c.f.][]{Mitchell_1980,Zador_2019}. \cite{hermann2020the} identified data augmentation as an important factor in increasing shape bias. Furthermore, \cite{feather2019metamers} generated synthetic DNN metamers: Stimuli that are physically different but have indistinguishable activity at a certain layer of a DNN. While metamers for the early layers were recognisable by human observers, the activity of later layer DNN metamers were not metameric for humans, pointing to a different representation of objects in DNNs and humans. Most DNNs do not use the same image features as humans do when recognising objects; but more recent models increasingly rely on object shapes and less on texture-like characteristics \citep{dehghani2023scaling}.

	\subsection{Susceptibility to adversarial attacks}
	\label{adversarial}
	\paragraph*{What are adversarial examples, and why are they problematic?}
	In a nutshell, \cite{szegedy2013intriguing} showed that for standard DNNs, \emph{any} image of category A can be made to be misclassified to belong to \emph{any} arbitary category B through a tiny perturbation, often so small that it is invisible to the human eye. Despite a decade of enormous research efforts to make DNNs robust against adversarial attacks, no principled defense mechanism has been found to date. The current best defense is brute-force \emph{adversarial training} \citep{madry2017towards}, which increases the perturbation needed to fool a DNN (an approach that may or may not achieve human-level robustness in the future). Adversarial examples are arguably one of the most pressing open problems of deep learning research, and researchers have started to ask whether adversarial examples exist for human perception, too. Typical methods to find adversarials for DNNs cannot be applied to human perception which is stochastic, sequence dependent and does not provide gradients. Despite those practical challenges, answering the question of whether there are adversarial examples for humans is highly relevant to the debate around DNNs as models of human visual perception \citep[e.g.][]{dujmovic2020adversarial}, since this would either indicate strong similarities or strong differences.
	
	\definecolor{box.colbacktitle}{RGB}{156, 174, 211}
	\definecolor{box.colback}{RGB}{217, 222, 238}
	\begin{figure}[t]
		\small
		\begin{tcolorbox}[width=\linewidth,colback={box.colback},
			title={\textbf{MYTH: HUMANS SUFFER FROM ADVERSARIAL VULNERABILITY}},
			colbacktitle=box.colbacktitle,coltitle=black]
			As explained in detail in section \ref{adversarial} there are no known adversarial examples for the human visual system: At least not adversarial examples according to the proper definition used in machine learning. Of course the visual system---like DNNs---is not error free: It can be shown to make wrong classifications if faced with hard images, ambiguous images or visual illusions. However, such images are not adversarial images---they are hard, ambiguous or visual-illusion images. All three kinds of images and their influence on human recognition are certainly worthy of study. But progress in vision science is hindered by confusing such images with (proper) adversarial images and, even worse, claiming that there are deep similarities between deep neural networks and human vision because both allegedly suffer from adversarial vulnerability. The opposite is true: Precisely because only DNNs are known to be vulnerable to adversarial examples, there are substantial and important differences between how DNNs and the human visual system process visual information.
		\end{tcolorbox}
	\end{figure}

	\paragraph*{The crocodile conjecture: Humans don't suffer from adversarial examples.}
	Do adversarial examples for humans exist? Strictly speaking the answer is unknown, but we believe it to be exceedingly unlikely that proper adversarial examples exist for humans. Unfortunately, while adversarial examples do have a precise definition, other types of images have in recent years been described in some way or another as ``adversarial''. Here we first precisely define what a convincing adversarial example for humans would be, applying the same definition that is used to define machine adversarials for a deep neural network $f$. Starting from an arbitrary image $i$ with ground truth label $l$ for which $f(i) = l$ holds (i.e.\ the original image is correctly classified by the model), an untargeted $\epsilon$-adversarial image is an image $i+\delta$ such that $f(i+\delta) \neq l$ and $||\delta||_p \leq \epsilon$. This adversarial example is \emph{untargeted} since the perturbation $\delta$ (which is small according to some $L_p$ norm, typically $p \in \{0,1,2,\infty\}$) just needs to fool the model into classifying $i+\delta$ as any class except for the original one. In the case of a \emph{targeted} adversarial, on the other hand, a target class $l'$ with $l' \neq l$ is chosen beforehand, and then any image $i+\delta$ such that $f(i+\delta) = l'$ is called a targeted adversarial example (subject to the same constraints as above, i.e.\ $f(i)=l$ and $||\delta||_p \leq \epsilon$).

	\newcommand{\subfigwidth}{0.23\linewidth}
	\begin{figure}
		\begin{subfigure}{\subfigwidth}
			\centering
			\textbf{banana (original)\vspace{0.1cm}}
			\includegraphics[width=\linewidth]{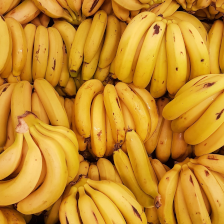}
		\end{subfigure}\hfill
		\begin{subfigure}{\subfigwidth}
			\centering
			\textbf{baseball\vspace{0.2cm}}
			\includegraphics[width=\linewidth]{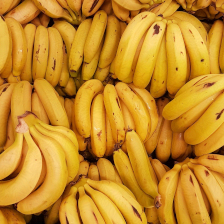}
		\end{subfigure}\hfill
		\begin{subfigure}{\subfigwidth}
			\centering
			\textbf{power drill\vspace{0.2cm}}
			\includegraphics[width=\linewidth]{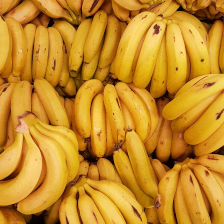}
		\end{subfigure}\hfill
		\begin{subfigure}{\subfigwidth}
			\centering
			\textbf{African crocodile\vspace{0.2cm}}
			\includegraphics[width=\linewidth]{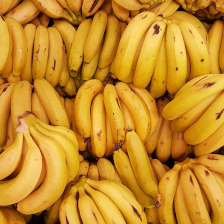}
		\end{subfigure}\hfill
		\caption{The crocodile conjecture: there are no adversarial examples for humans, i.e.\ it is impossible to make humans classify the bananas in the left image as a crocodile with a tiny perturbation. The original banana image (left) is adversarially modified (by adding tiny perturbations) such that a ResNet-50 then classifies the resulting perturbed adversarial images as either baseball, power drill or African crocodile (depending on the perturbation). Results based on the pre-trained torchvision \citep{marcel2010torchvision} ResNet-50 implementation and the targeted $L_{\infty}$ PGD attack with $\epsilon=0.01$ implemented by Foolbox \citep{rauber2017foolbox}. \textit{Image source for original banana image (left): Photo by \href{https://unsplash.com/@rodreis}{Rodrigo dos Reis} on \href{https://unsplash.com/s/photos/bananas?utm_source=unsplash&utm_medium=referral&utm_content=creditCopyText}{Unsplash}, licensed under the \href{https://unsplash.com/license}{Unsplash license}.}}
		\label{fig:adversarial_grid}
	\end{figure}
	
	If we transfer this to human visual perception, this would mean that if humans are indeed susceptible to $\epsilon$-targeted adversarial examples, then for an arbitrary image $i$ such as the bananas in Figure~\ref{fig:adversarial_grid} there would be a small perturbation $\delta$ ($||\delta||_p \leq \epsilon$) such that a human observer would classify $i+\delta$ as an arbitrary but pre-specified other class, such as a crocodile. If $\epsilon$ is large enough, this is trivially possible: We simply replace the banana image by a crocodile image. Hence, the crucial aspect is that the perturbation bounded by $\epsilon$ needs to be small. A standard ResNet-50 model, for example, can be fooled into classifying the banana image into a baseball/power drill/crocodile image with $\epsilon=0.01$ (using the $L_{\infty}$ norm and the [0, 1] image range), as shown in Figure~\ref{fig:adversarial_grid}. This means that while every pixel may be changed, the maximum allowed perturbation is 0.01/1 or 1\% for each pixel. For humans, we believe it is impossible to find a perturbation $\delta$ subject to $||\delta||_{\infty} \leq 0.01$ that yields a ``crocodile'' classification for the banana image. We call this the \emph{crocodile conjecture} (no adversarial examples for humans).\footnote{We would love to be proven wrong; thus for anyone attempting to achieve this, a convincing demonstration would consist of a forced choice paradigm where the choices are either banana or crocodile, with viewing times of at least 100--200 ms, foveal presentation, $\epsilon=0.01$ and $p=\infty$.} Given the many other types of images that have been termed ``adversarial'', we believe it is important to be precise when considering what a convincing adversarial example for human visual perception would constitute. In the following, we will attempt to explain what adversarial examples are \emph{not}: neither hard images, nor ambiguous images, nor visual illusions.
	
	\paragraph*{What adversarial examples are not: hard images.}
	To err is human, and making errors is a part of perception, whether biological or artificial.  A popular image dataset refers to natural, unmodified images that are particularly hard, i.e.\ images on which making errors is likely, as ``Natural Adversarial Examples'' \citep{hendrycks2021natural}. While the dataset is great, the name is not: these hard images have nothing to do with adversarial examples.
	
	\paragraph*{What adversarial examples are not: ambiguous images.}
	Just as the existence of hard images is normal, so is the existence of ambiguous images. Examples include the famous rabbit-duck image \citep{jastrow1899mind}, hybrid images consisting of conflicting high- and low spatial frequencies \citep{oliva2006hybrid} or an image that resembles both a cat and a dog as in \cite{elsayed2018adversarial} (Figure~1 in arXiv v1 and v2, Figure Supp.~1 in arXiv v3), or multistable images \citep[e.g.][]{Safavi-Dayan_2022}. Convincing adversarial examples are not tied to inherently ambiguous images---in fact, adversarial examples are problematic for DNNs precisely because they can be found for arbitrary images.
	
	\paragraph*{What adversarial examples are not: visual illusions.}
	While the precise definition of visual illusions is subject to debate \citep{todorovic2020visual}, it commonly involves a discrepancy between perception and ``reality''. According to this definition, adversarial examples---if humans were susceptible to them!---might be considered a new type of visual illusion. Importantly, however, the reverse is not true: None of the visual illusions we know are adversarial examples. Like ambiguous images, known visual illusions are non-general; illusions are very particular stimuli or images like the particular configuration of shapes in case of the Kanisza triangle, a particular configuration of luminances (reflectances) in case of lightness and brightness illusions, or a particular viewing angle as in the case of the Necker cube. No (known) visual illusion matches the definition of an adversarial example.
	
	\paragraph*{Relevance of machine adversarials for humans.}
	While no proper adversarials for humans are known, a number of studies investigated whether machine adversarials contain patterns that have relevance to humans, too. \cite{zhou2019humans} showed that humans can sometimes anticipate the predictions of deep learning models to adversarial patterns---which is very different from being susceptible to adversarial patterns (also see \cite{dujmovic2020adversarial} for control experiments). Furthermore, \cite{elsayed2018adversarial} showed that human classification decisions at short presentation times (63--71 ms)\footnote{In addition images were not shown in the central fovea (approx.\ 2 degrees of visual angle) but were rather large, extending to 14.2 degrees of visual angle and thus into the periphery.} are influenced, to a certain degree, by intermediate-size perturbations that are adversarial to deep learning models. Thus intermediate-size (rather than small-size) adversarial perturbations are more than just arbitrary noise to humans. Unfortunately, the study is often erroneously interpreted as showing that there are adversarial examples for humans. The study's design choices purposefully target a regime where humans are highly prone to making errors (baseline classification performance for the choice between two classes is at around 75\%, where 50\% is chance level), and the study only used semantically related classes (dog/cat, cabbage/broccoli, spider/snake), not arbitrary ones. These are sensible choices for studying small effects under extreme conditions, but this is different from human core object recognition \citep{dicarlo2012does}. \cite{elsayed2018adversarial} studied edge cases, using a paradigm that cannot test general perception for arbitrary classes (e.g. banana/crocodile) under conditions where the original image can be categorized with high accuracy.
	
	Taken together, we believe that humans don't have adversarials that can turn any arbitrary image (such as a banana) into an arbitrary other category (such as a crocodile) with only a tiny perturbation---the crocodile conjecture. Humans rely heavily on shape when recognising objects, and object shape and boundaries cannot be significantly changed using tiny perturbations only. Adversarial examples are a profound problem for DNNs, but not for humans.

	\definecolor{box.colbacktitle}{RGB}{156, 174, 211}
	\definecolor{box.colback}{RGB}{217, 222, 238}
	\begin{figure}[t]
		\small
		\begin{tcolorbox}[width=\linewidth,colback={box.colback},
			title={\textbf{MYTH: CERTAIN TASKS CANNOT EVER BE SOLVED WITH DEEP NEURAL NETWORKS}},
			colbacktitle=box.colbacktitle,coltitle=black]
			Even though DNNs are getting better and better, there are still many tasks on which humans outperform them. It is therefore tempting to investigate whether certain tasks \emph{cannot ever} be solved with deep neural networks. We believe that such efforts are unlikely to succeed in general using empirical investigations: First of all, the broader class of deep neural networks is theoretically capable of solving \emph{any} task, i.e.\ representing any conceivable input-output mapping \citep{hornik1989multilayer}. Secondly, training a specific network (or a handful of them) can never serve as a proof of nonexistence. This error is not infrequently witnessed in vision science when claims are made that a certain task cannot be solved by DNNs because AlexNet or ResNet-50 or some other model cannot solve the task (given a certain dataset, training regime, objective function, and the researcher's technical prowess). When investigating the question of whether a task cannot be solved by specific networks through a set of experiments, the results need to be contextualised, highlighting the necessarily restricted set of explorations and thus the necessarily restricted set of conclusions (context which more often than not appears to be challenging to fit into a crisp paper title, abstract or general summary). In contrast to empirical investigations, theoretical proofs have the beauty that they can sometimes make very general statements about a class of models---such as, for instance, the proof that DNNs with ReLU activation functions are almost always overconfident on (some) errors~\citep{Hein-etal_2019a}.
			Thus there is an important asymmetry: Successfully training a specific DNN on some task can serve at least as a proof-of-concept. Failure to train a specific DNN on some task has limited implications; all it shows is that the specific DNN, with its dataset, training regime, objective function etc.\ failed---a different DNN architecture, dataset, training regime or objective function may have succeeded unless there is theoretical proof why a task is beyond DNNs.
		\end{tcolorbox}
	\end{figure}
	
	\definecolor{box.colbacktitle}{RGB}{156, 174, 211}
	\definecolor{box.colback}{RGB}{217, 222, 238}
	\begin{figure}[t]
		\small
		\begin{tcolorbox}[width=\linewidth,colback={box.colback},
			title={\textbf{MYTH: MORE (OF THE SAME) DATA IS ALL WE NEED}},
			colbacktitle=box.colbacktitle,coltitle=black]
			DNNs typically require in excess of a million images; one of the latest SWAG models by \citet{singh2022revisiting} was trained on 3.6 billion images. While vision scientists recognise the crucial role of carefully controlled stimuli in experiments, large sets of images are impossible to curate on an image-by-image basis, however, and image datasets are harvested from the internet---with the hope that there is enough variation in scenes, objects, illuminations, pose and viewpoint variations etc.\ such that there are, first, no systematic biases and, second, appropriately distributed (recognition) difficulties of the individual images.\\
			
			Neither belief may be warranted, however: \cite{meding2021trivial} found image difficulty levels in ImageNet to be highly unbalanced, containing far too many trivial (and impossible) images. Furthermore, systematic biases in computer vision datasets typically do not disappear with dataset scale, and ML models often exploit those biases. Examples are the spectral bias resulting from large aperture ``portrait-style'' photos with shallow depth-of-field in ``rapid animal detection'' \citep{wichmann2010animal} or other cases of dataset bias \citep{torralba2011unbiased}. Exploiting shortcuts in the data \citep{geirhos2020shortcut} can lead to various generalisation failures. One striking example is the 10--12\% classification accuracy drop of DNNs when tested on a new ``ImageNet 2.0'' dataset which \citet{Recht-etal_2019} created faithfully mimicking the original curation process: ``This suggests that the accuracy scores of even the best image classifiers are still highly sensitive to minutiae of the data cleaning process. $\dots$ It also shows that current classifiers still do not generalize reliably even in the benign environment of a carefully controlled reproducibility experiment'' \cite[p.~1]{Recht-etal_2019}.\\
			
			Machine learning methods typically assume training and test sets to be independent and identically distributed (iid). For real-world natural images, what exactly are iid-images is still an unresolved issue. Is a photograph of the same scene during a different season an ``independent'' image? One taken under different lighting?  Or already one taken with the camera moved a little to one side or up or down? Are photos of a group of people taken from a different vantage point with different facial expressions ``independent'' images?  Progress towards understanding these issues will likely help making DNNs behave more human like.
		\end{tcolorbox}
	\end{figure}

	\definecolor{box.colbacktitle}{RGB}{156, 174, 211}
	\definecolor{box.colback}{RGB}{217, 222, 238}
	\begin{figure}[t]
		\small
		\begin{tcolorbox}[width=\linewidth,colback={box.colback},
			title={\textbf{MYTH: RECURRENCE IS NECESSARY TO SOLVE CERTAIN TASKS}},
			colbacktitle=box.colbacktitle,coltitle=black]
			Given the important behavioural differences between DNNs and biological vision, one may wonder where these differences originate from. One clear difference between brains and standard DNNs is that brains have recurrent connections, unlike standard feedforward DNNs. In line with this, a number of papers argue for the importance of recurrent processing for both biological and artificial systems \citep[e.g.][]{serre2019deep, kietzmann2019recurrence, kubilius2019brain, kreiman2020beyond, vanBergen-Kriegeskorte_2020}.
			Going beyond importance, however, sometimes the argument is made that recurrent DNNs are necessary to solve certain---challenging---tasks. In contrast, we do not think that recurrence is the key missing ingredient since any algorithm that can be implemented by a recurrent DNN can also be implemented by a computationally equivalent feedforward DNN.\\
			
			Since recurrent networks ``once unfolded in time [...], can be seen as very deep feedforward networks in which all the layers share the same weights'' \citep[p.~442]{lecun2015deep}, there is no difference whatsoever between a finite time recurrent network and its unrolled feedforward counterpart in terms of what the model can compute. Any \emph{finite time} recurrent network can be represented by a computationally equivalent finite depth feedforward network (e.g.\ via unrolling), and any \emph{infinite time} recurrent network can be represented by a computationally equivalent infinitely deep feedforward network (keeping in mind that neither infinite time nor infinite depth would be particularly biologically plausible). Hence, there is no task that can only be solved with recurrence. For certain tasks, reusing weights may be useful---but this can be equivalently achieved by feedforward networks with weight sharing and by recurrent networks. In fact, in current DNN software libraries most recurrent networks are trained as unfolded or unrolled feedforward networks with an algorithm called, tellingly, \emph{backpropagation through time}---an implementational aspect well known by the researchers cited above and in the machine learning community in general. Any claim that recurrent networks have a computational advantage over feedforward networks---that is: can solve tasks that feedforward networks cannot solve---thus appears mistaken.\\
			
			We believe that much of the debate around recurrence can be clarified when distinguishing between \emph{computation} and \emph{implementation}. As described above, any recurrent network can be unfolded into a (potentially very or even impractically deep) computationally equivalent feedforward network. At the implementational level, however, there are clear and important differences even between computationally equivalent recurrent and feedforward networks. For instance, artificial recurrent networks need less space to fit into memory, and biological ones need fewer neurons to fit into a brain, and have obvious advantages in terms of energy efficiency and flexibility of processing. Whether one cares more about implementation or computation depends on the investigated question---but what is important is to distinguish between the two.
		\end{tcolorbox}
	\end{figure}

	\section{DISCUSSION AND OPEN ISSUES}
	\label{disussion}
	
	\paragraph*{Status quo: Not adequate, but promising.}
	As described in Section~\ref{assessmentDNN}, in spite of their excellent prediction performance on standard image datasets like ImageNet, current DNNs see the world differently from human observers. We have reviewed evidence that DNNs still lack robustness to changes in object pose (Section~\ref{3Dviewpoints}) and image distortions (Section~\ref{imageDistortions}), make non-human-like errors as assessed by error consistency (last paragraph of Section~\ref{imageDistortions}), exhibit a lack of human-level shape bias (Section~\ref{imageFeatures}) and show a worrying and non-human susceptibility to adversarial images (Section~\ref{adversarial}). These behavioural differences can be exemplified in the following thought experiment: We could generate a dataset containing only adversarial images and images with slightly different 3D viewpoints. While human classification performance would be largely unaffected, DNN performance, on the other hand, could be driven to 0\% correct. Conversely, given the low error-consistency between DNNs and humans, we could select a subset of non-adversarially distorted images for which human performance is consistently poor but DNN performance excellent. In effect we could (almost) create a double-dissociation between DNNs and human observers in terms of core object recognition performance---impossible if both DNNs and humans recognised images similarly.
	These profound behavioural differences indicate that current DNNs are not yet adequate behavioural models of human core object recognition. At the same time, we would like to stress that assessing the (in)adequacy of DNNs as models of human core object recognition behaviour can only be a snapshot in time---it is true as of today. This does not mean that DNNs are forever, or for theoretical reasons, incapable of becoming adequate models of human core object recognition. In fact, for all of the current challenges described in Section~\ref{assessmentDNN}, the tremendous pace of progress in deep learning will very likely lead to improvements. Indeed, DNNs struggle with certain distortions---but even so, more recent models have come a long way from AlexNet performance and even show more human-aligned error patterns \citep{geirhosPartialSuccessClosing2021}. Indeed, DNNs are susceptible to adversarial perturbations---but even so, adversarial training keeps improving robustness \citep{madry2017towards, croce2020robustbench}. Not always are the same models successful along all desired dimensions---for instance, adversarially trained models perform particularly poorly when exposed to rotated or low-contrast images \citep{geirhosPartialSuccessClosing2021}---but the pace of progress overall is substantial and we are keen to see how the next generation of deep learning models might become more behaviourally faithful.
	
	\paragraph*{Excitement vs.\ disappointment.}
	Deep learning is often described as a revolution, and just like with any revolution there is tremendous excitement and simultaneously profound disappointment. Vision science is no different: on the one hand, the ability of DNNs to fit neural data has led to the assessment that ``deep hierarchical neural networks are beginning to transform neuroscientists’ ability to produce quantitatively accurate computational models of the sensory systems'' \citep[][p.~364]{Yamins2016a} or claims beyond core object recognition that ``[d]eep neural networks provide the current best models of visual information processing in the primate brain'' \citep[p.~1]{Mehrer-etal_2021}. On the other hand, behavioural shortcomings have also led to the assessment that there are ``deep problems with neural network models of human vision'' \citep[][p.~1]{Bowers-etal_2022}. We believe that both of those perspectives are understandable, and that some---not all---of these seemingly contradictory accounts can be reconciled through greater clarity about the goal of a particular model. Model quality is not a one-dimensional construct: some models are good in some regards and poor in others. In particular, DNNs are the best \emph{predictive} models in the history of core object recognition models---on computer vision benchmarks. At the same time, they struggle to show human-like performance on more challenging, human vision inspired performance tests like 3D viewpoint variation or distortion robustness and have low error consistency with humans; furthermore, currently DNNs are also probably among the most inscrutable models, which constitutes a challenge towards the modelling goal of \emph{explanation}.
	
	\paragraph*{Future direction: Vision science for deep learning.}
	In Section~\ref{tools}, we talked about the deep learning as tools in science and vision science. Later, we have seen numerous examples of the value of vision science for deep learning: to understand how they work, where they fail, and how they see the world. Psychophysical studies have revealed current limitations of DNNs that are of interest not only to those interested in building better behavioural models, but better deep learning models in general. Careful and fair comparisons \citep[c.f.][]{Funke-etal_2021} are the hallmark of vision science, and the field with its strong scientific foundation has much to offer to deep learning, which is still a predominantly engineering-driven area. In return, we may hope to develop better behavioural models of human visual perception, benefiting from rapid advances in deep learning and the field's engineering ingenuity \citep{ma2020neural, Peters-Kriegeskorte_2021b}.
	Interesting developments in the direction of better vision science inspired methods for scrutinising deep learning include generating \emph{controversial stimuli} \citep{Golan-etal_2020a} to distinguish between candidate models \citep[related to the idea of maximum differentiation competition, or MAD, by][]{Wang_2008}, using crowding measures to assess the importance of local vs.\ global features \citep{Doerig-etal_2020a}, and using a comparative biology approach when comparing human and machine visual perception \citep{Lonnqvist-etal_2021a}, just to name a few. We are convinced that deep learning can benefit from vision science just as much as vision science can benefit from deep learning methods.
	This will be particularly true if we move beyond core object recognition to visual perception in general. Some of the challenges ahead are well articulated in \cite{Lake_2017} and \cite{Depeweg_2018}; a more sceptical position regarding DNNs as (only) models of perception and cognition is provided by \cite{marcus2018deep}.

	\paragraph*{Future direction: Understanding inductive biases.}
	In machine learning the necessity to make assumptions in order to learn anything useful has been formalised by the \emph{no free lunch theorem} \citep{wolpert1997no}. But understanding the inductive bias of a particular deep learning model---i.e., the set of assumptions that the model makes ahead of being exposed to data---is often incredibly challenging, and there are many complex interactions between, for instance, model architecture and dataset. In spite of those challenges, we believe that it will be very important to improve our understanding of the assumptions that perceptual systems make. Humans, for instance, are far from being a tabula rasa at birth \citep{Zador_2019}, we have a highly structured brain with appropriate inductive bias to help us learn rapidly and robustly from comparatively little data---while deep neural networks, at the moment, still sometimes require more data than a human can possibly be exposed to during their lifetime \citep{huber2022developmental}.
	There is much that remains to be understood: How do models and tasks interact with datasets (also see box on MYTH: MORE (OF THE SAME) DATA IS ALL WE NEED)? What are the inductive biases of biologically-inspired filters \citep{dapello2020simulating,Evans-etal_2022}? We witness that ``recent work in AI [...] increasingly favoured computational architectures (for example, graph nets and tranformers) that implement an inductive bias towards relational, compositional processing'' \citep[][p.~1263]{Piloto-etal_2022}, while standard, simpler convolutional networks are becoming less and less relevant. In a similar direction, \cite{sabour2017dynamic} introduced capsule networks arguing that DNNs should have an inductive bias for explicit representation of geometric relationships of objects, which helps in order to be viewpoint invariant \citep{kosiorek2019stacked}. As a final question, may we require an explicit mid-level representation, the \emph{pre-semantic} experience of the world \citep{Nakayama_1995,Anderson_2020a}? Currently DNNs are trained \emph{end-to-end}, from pixels to semantics in one model using one objective function. Should we explicitly train DNNs on psychologically inspired mid-level representations, and then go from them to semantics?
	Building more adequate models of visual perception will likely require increasing attention to the various explicit and implicit assumptions, or inductive biases, made by different models.

	\section{CONCLUSION}
	A decade ago no researcher in vision science would have foreseen the phenomenal progress made by machine learning researchers in the field of neural networks. DNNs are spectacularly successful tools for (vision) science but also  promising (statistical) models of core object recognition in terms of their prediction performance on standard computer vision image datasets. However, it is also fair to say that few researchers foresaw the substantial problems remaining for DNNs despite their excellent prediction performance on standard computer vision image datasets: their lack of robustness to object pose and image distortions, their non-human-like errors as assessed by error consistency, their as of now ill-understood dependence on the minutiae of the training images, their lack of human-like shape encoding as well as their susceptibility to adversarial images. Last but not least we are still lacking reliable tools to turn well-predicting but complex and non-transparent DNNs into human-understandable explanations---a desideratum of a scientific model beyond purely statistical models. We argue, thus, that as of today DNNs should be regarded as promising---but not yet adequate---models of human core object recognition performance.
	
	\small{
		\paragraph*{Acknowledgments}
		We would like to thank, in alphabetic order, Guillermo Aguilar, Bart Anderson, Blair Bilodeau, Wieland Brendel, Stéphane Deny, Roland Fleming, Tim Kietzmann, Thomas Klein, Been Kim, Pang Wei Koh, Simon Kornblith, David-Elias Künstle, Marianne Maertens, Sascha Meyen, Lisa Schut, Heiko Schütt, Kate Storrs and Uli Wannek for helpful discussions and/or valuable feedback on aspects of the manuscript, and Roland Zimmermann for foolbox advice. Furthermore, the authors are particularly indebted to Frank Jäkel and his numerous critical and insightful comments on previous versions of our manuscript. All opinions expressed in this article are our own and are not necessarily shared by any of the colleagues we thank here.\\
		Felix Wichmann is a member of the Machine Learning Cluster of Excellence, funded by the Deutsche Forschungsgemeinschaft (DFG, German Research Foundation) under Germany’s Excellence Strategy – EXC number 2064/1 – Project number 390727645.
	}
	
	\begin{small}
		\bibliographystyle{plainnat}
		\bibliography{refs}

\begin{thebibliography}{131}
\providecommand{\natexlab}[1]{#1}
\providecommand{\url}[1]{\texttt{#1}}
\expandafter\ifx\csname urlstyle\endcsname\relax
  \providecommand{\doi}[1]{doi: #1}\else
  \providecommand{\doi}{doi: \begingroup \urlstyle{rm}\Url}\fi

\bibitem[Abbas and Deny(2022)]{abbas2022progress}
Amro Abbas and St{\'e}phane Deny.
\newblock Progress and limitations of deep networks to recognize objects in
  unusual poses.
\newblock \emph{arXiv:2207.08034}, 2022.

\bibitem[Adebayo et~al.(2018)Adebayo, Gilmer, Muelly, Goodfellow, Hardt, and
  Kim]{adebayo2018sanity}
Julius Adebayo, Justin Gilmer, Michael Muelly, Ian Goodfellow, Moritz Hardt,
  and Been Kim.
\newblock Sanity checks for saliency maps.
\newblock In \emph{{Advances in Neural Information Processing Systems}, 32},
  pages 9505--9515, 2018.

\bibitem[Alcorn et~al.(2019)Alcorn, Li, Gong, Wang, Mai, Ku, and
  Nguyen]{alcorn2019strike}
Michael~A Alcorn, Qi~Li, Zhitao Gong, Chengfei Wang, Long Mai, Wei-Shinn Ku,
  and Anh Nguyen.
\newblock Strike (with) a pose: Neural networks are easily fooled by strange
  poses of familiar objects.
\newblock In \emph{{Proceedings of the IEEE Conference on Computer Vision and
  Pattern Recognition}}, 2019.

\bibitem[Anderson(2020)]{Anderson_2020a}
Barton~L. Anderson.
\newblock Mid-level vision.
\newblock \emph{Current Biology}, 30\penalty0 (3):\penalty0 R105--R109,
  February 2020.
\newblock ISSN 0960-9822.
\newblock \doi{10.1016/j.cub.2019.11.088}.

\bibitem[Baker et~al.(2018)Baker, Lu, Erlikhman, and Kellman]{baker2018deep}
Nicholas Baker, Hongjing Lu, Gennady Erlikhman, and Philip~J Kellman.
\newblock Deep convolutional networks do not classify based on global object
  shape.
\newblock \emph{{PLoS Computational Biology}}, 14\penalty0 (12):\penalty0
  e1006613, 2018.

\bibitem[Berardino et~al.(2017)Berardino, Laparra, Ball{\'e}, and
  Simoncelli]{berardino2017eigen}
Alexander Berardino, Valero Laparra, Johannes Ball{\'e}, and Eero Simoncelli.
\newblock Eigen-distortions of hierarchical representations.
\newblock \emph{{Advances in Neural Information Processing Systems}}, 30, 2017.

\bibitem[Biederman(1987)]{Biederman_1987}
Irving Biederman.
\newblock Recognition-by-components: {{A}} theory of human image understanding.
\newblock \emph{Psychological Review}, 94\penalty0 (2):\penalty0 115--147,
  1987.
\newblock ISSN 0033-295X.

\bibitem[Boelts et~al.(2022)Boelts, Lueckmann, Gao, and
  Macke]{Boelts-etal_2022}
Jan Boelts, Jan-Matthis Lueckmann, Richard Gao, and Jakob~H Macke.
\newblock Flexible and efficient simulation-based inference for models of
  decision-making.
\newblock \emph{eLife}, 11:\penalty0 e77220, July 2022.
\newblock ISSN 2050-084X.
\newblock \doi{10.7554/eLife.77220}.

\bibitem[Borowski et~al.(2021)Borowski, Zimmermann, Schepers, Geirhos, Wallis,
  Bethge, and Brendel]{Borowski-etal_2021}
Judy Borowski, Roland~S. Zimmermann, Judith Schepers, Robert Geirhos, Thomas
  S.~A. Wallis, Matthias Bethge, and Wieland Brendel.
\newblock Exemplary {Natural} {Images} {Explain} {CNN} {Activations} {Better}
  than {State}-of-the-{Art} {Feature} {Visualization}.
\newblock In \emph{International Conference on Learning Representations}, May
  2021.
\newblock \doi{10.48550/arXiv.2010.12606}.
\newblock URL \url{http://arxiv.org/abs/2010.12606}.

\bibitem[Bowers et~al.(2022)Bowers, Malhotra, Dujmovi{\'c}, Montero, Tsvetkov,
  Biscione, Puebla, Adolfi, Hummel, Heaton, Evans, Mitchell, and
  Blything]{Bowers-etal_2022}
Jeffrey~S. Bowers, Gaurav Malhotra, Marin Dujmovi{\'c}, Milton~Llera Montero,
  Christian Tsvetkov, Valerio Biscione, Guillermo Puebla, Federico~G. Adolfi,
  John Hummel, Rachel~Flood Heaton, Benjamin Evans, Jeff Mitchell, and Ryan
  Blything.
\newblock Deep problems with neural network models of human vision, April 2022.

\bibitem[Box(1976)]{Box_1976}
George E~P Box.
\newblock Science and {{Statistics}}.
\newblock \emph{Journal of the American Statistical Association}, 71\penalty0
  (356):\penalty0 791--799, 1976.
\newblock ISSN 0162-1459.

\bibitem[Brendel and Bethge(2019)]{brendel2019approximating}
Wieland Brendel and Matthias Bethge.
\newblock Approximating {CNNs} with bag-of-local-features models works
  surprisingly well on {ImageNet}.
\newblock In \emph{{International Conference on Learning Representations}},
  2019.

\bibitem[Brick et~al.(2021)Brick, Hood, Ekroll, and
  {de-Wit}]{brickIllusoryEssencesBias2021}
C.~Brick, B.~Hood, V.~Ekroll, and L.~{de-Wit}.
\newblock Illusory {{Essences}}: {{A Bias Holding Back Theorizing}} in
  {{Psychological Science}}.
\newblock \emph{Perspectives on Psychological Science}, page 1745691621991838,
  July 2021.
\newblock ISSN 1745-6916.
\newblock \doi{10.1177/1745691621991838}.

\bibitem[Cajal(1906)]{Cajal_1906}
Santiago Ramón~Y Cajal.
\newblock The structure and connexions of neurons.
\newblock \emph{Nobel Prize Lecture}, pages 220--253, 1906.

\bibitem[Chung and Abbott(2021)]{Chung-Abbott_2021}
SueYeon Chung and L.~F. Abbott.
\newblock neural population geometry: {{An}} approach for understanding
  biological and artificial neural networks.
\newblock \emph{Current Opinion in Neurobiology}, 70:\penalty0 137--144,
  October 2021.
\newblock ISSN 0959-4388.
\newblock \doi{10.1016/j.conb.2021.10.010}.

\bibitem[Cichy and Kaiser(2019)]{Cichy_2019}
Radoslaw~M. Cichy and Daniel Kaiser.
\newblock Deep {{neural networks}} as {{scientific models}}.
\newblock \emph{{Trends in Cognitive Sciences}}, 2019.
\newblock ISSN 1364-6613.

\bibitem[Cohen et~al.(2020)Cohen, Chung, Lee, and Sompolinsky]{Cohen-etal_2020}
Uri Cohen, SueYeon Chung, Daniel~D. Lee, and Haim Sompolinsky.
\newblock Separability and geometry of object manifolds in deep neural
  networks.
\newblock \emph{{Nature Communications}}, 11\penalty0 (1):\penalty0 746,
  December 2020.
\newblock ISSN 2041-1723.
\newblock \doi{10.1038/s41467-020-14578-5}.

\bibitem[Croce et~al.(2020)Croce, Andriushchenko, Sehwag, Debenedetti,
  Flammarion, Chiang, Mittal, and Hein]{croce2020robustbench}
Francesco Croce, Maksym Andriushchenko, Vikash Sehwag, Edoardo Debenedetti,
  Nicolas Flammarion, Mung Chiang, Prateek Mittal, and Matthias Hein.
\newblock Robustbench: a standardized adversarial robustness benchmark.
\newblock \emph{arXiv:2010.09670}, 2020.

\bibitem[Dapello et~al.(2020)Dapello, Marques, Schrimpf, Geiger, Cox, and
  DiCarlo]{dapello2020simulating}
Joel Dapello, Tiago Marques, Martin Schrimpf, Franziska Geiger, David~D Cox,
  and James~J DiCarlo.
\newblock Simulating a primary visual cortex at the front of {CNNs} improves
  robustness to image perturbations.
\newblock \emph{bioRxiv}, 2020.

\bibitem[Dax et~al.(2021)Dax, Green, Gair, Macke, Buonanno, and
  Sch{\"o}lkopf]{Dax-etal_2021}
Maximilian Dax, Stephen~R. Green, Jonathan Gair, Jakob~H. Macke, Alessandra
  Buonanno, and Bernhard Sch{\"o}lkopf.
\newblock Real-{{Time Gravitational Wave Science}} with {{Neural Posterior
  Estimation}}.
\newblock \emph{Physical Review Letters}, 127\penalty0 (24):\penalty0 241103,
  December 2021.
\newblock ISSN 0031-9007, 1079-7114.
\newblock \doi{10.1103/PhysRevLett.127.241103}.

\bibitem[Dehghani et~al.(2023)Dehghani, Djolonga, Mustafa, Padlewski, Heek,
  Gilmer, Steiner, Caron, Geirhos, Alabdulmohsin, et~al.]{dehghani2023scaling}
Mostafa Dehghani, Josip Djolonga, Basil Mustafa, Piotr Padlewski, Jonathan
  Heek, Justin Gilmer, Andreas Steiner, Mathilde Caron, Robert Geirhos, Ibrahim
  Alabdulmohsin, et~al.
\newblock Scaling vision transformers to 22 billion parameters.
\newblock In \emph{{International Conference on Machine Learning}}, 2023.

\bibitem[Depeweg et~al.(2018)Depeweg, Rothkopf, and Jäkel]{Depeweg_2018}
Stefan Depeweg, Constantin~A Rothkopf, and Frank Jäkel.
\newblock Solving {Bongard} {problems} with a {visual} {language} and
  {pragmatic} {reasoning}.
\newblock \emph{arXiv}, 1804.04452, 2018.

\bibitem[DiCarlo et~al.(2012)DiCarlo, Zoccolan, and Rust]{dicarlo2012does}
James~J DiCarlo, Davide Zoccolan, and Nicole~C Rust.
\newblock How does the brain solve visual object recognition?
\newblock \emph{Neuron}, 73\penalty0 (3):\penalty0 415--434, 2012.

\bibitem[Doerig et~al.(2020)Doerig, Schmittwilken, Sayim, Manassi, and
  Herzog]{Doerig-etal_2020a}
Adrien Doerig, Lynn Schmittwilken, Bilge Sayim, Mauro Manassi, and Michael~H.
  Herzog.
\newblock Capsule networks as recurrent models of grouping and segmentation.
\newblock \emph{{PLoS Computational Biology}}, 16\penalty0 (7):\penalty0
  e1008017, July 2020.
\newblock ISSN 1553-7358.
\newblock \doi{10.1371/journal.pcbi.1008017}.

\bibitem[Dong et~al.(2022)Dong, Ruan, Su, Kang, Wei, and Zhu]{Dong-etal_2022}
Yinpeng Dong, Shouwei Ruan, Hang Su, Caixin Kang, Xingxing Wei, and Jun Zhu.
\newblock Viewfool: evaluating the robustness of visual recognition to
  adversarial viewpoints.
\newblock \emph{Advances in Neural Information Processing Systems}, 36, October
  2022.
\newblock \doi{10.48550/arXiv.2210.03895}.

\bibitem[Dosovitskiy et~al.(2020)Dosovitskiy, Beyer, Kolesnikov, Weissenborn,
  Zhai, Unterthiner, Dehghani, Minderer, Heigold, Gelly,
  et~al.]{dosovitskiy2020image}
Alexey Dosovitskiy, Lucas Beyer, Alexander Kolesnikov, Dirk Weissenborn,
  Xiaohua Zhai, Thomas Unterthiner, Mostafa Dehghani, Matthias Minderer, Georg
  Heigold, Sylvain Gelly, et~al.
\newblock An image is worth 16x16 words: Transformers for image recognition at
  scale.
\newblock \emph{arXiv:2010.11929}, 2020.

\bibitem[Douglas and Martin(1991)]{Douglas_1991}
Rodney~J. Douglas and Kevan~A.C. Martin.
\newblock Opening the grey box.
\newblock \emph{Trends in Neurosciences}, 14\penalty0 (7):\penalty0 286--293,
  1991.
\newblock ISSN 0166-2236.

\bibitem[Dujmovi{\'c} et~al.(2020)Dujmovi{\'c}, Malhotra, and
  Bowers]{dujmovic2020adversarial}
Marin Dujmovi{\'c}, Gaurav Malhotra, and Jeffrey~S Bowers.
\newblock What do adversarial images tell us about human vision?
\newblock \emph{Elife}, 9:\penalty0 e55978, 2020.

\bibitem[Elsayed et~al.(2018)Elsayed, Shankar, Cheung, Papernot, Kurakin,
  Goodfellow, and Sohl-Dickstein]{elsayed2018adversarial}
Gamaleldin Elsayed, Shreya Shankar, Brian Cheung, Nicolas Papernot, Alexey
  Kurakin, Ian Goodfellow, and Jascha Sohl-Dickstein.
\newblock Adversarial examples that fool both computer vision and time-limited
  humans.
\newblock \emph{{Advances in Neural Information Processing Systems}}, 31, 2018.

\bibitem[Evans et~al.(2022)Evans, Malhotra, and Bowers]{Evans-etal_2022}
Benjamin~D. Evans, Gaurav Malhotra, and Jeffrey~S. Bowers.
\newblock Biological convolutions improve {DNN} robustness to noise and
  generalisation.
\newblock \emph{Neural Networks}, 148:\penalty0 96--110, April 2022.
\newblock ISSN 0893-6080.
\newblock \doi{10.1016/j.neunet.2021.12.005}.
\newblock URL
  \url{https://www.sciencedirect.com/science/article/pii/S0893608021004780}.

\bibitem[Fawzi et~al.(2022)Fawzi, Balog, Huang, Hubert, {Romera-Paredes},
  Barekatain, Novikov, R.~Ruiz, Schrittwieser, Swirszcz, Silver, Hassabis, and
  Kohli]{Fawzi-etal_2022}
Alhussein Fawzi, Matej Balog, Aja Huang, Thomas Hubert, Bernardino
  {Romera-Paredes}, Mohammadamin Barekatain, Alexander Novikov, Francisco~J.
  R.~Ruiz, Julian Schrittwieser, Grzegorz Swirszcz, David Silver, Demis
  Hassabis, and Pushmeet Kohli.
\newblock discovering faster matrix multiplication algorithms with
  reinforcement learning.
\newblock \emph{{Nature}}, 610\penalty0 (7930):\penalty0 47--53, October 2022.
\newblock ISSN 1476-4687.
\newblock \doi{10.1038/s41586-022-05172-4}.

\bibitem[Feather et~al.(2019)Feather, Durango, Gonzalez, and
  McDermott]{feather2019metamers}
Jenelle Feather, Alex Durango, Ray Gonzalez, and Josh McDermott.
\newblock Metamers of neural networks reveal divergence from human perceptual
  systems.
\newblock \emph{{Advances in Neural Information Processing Systems}}, 32, 2019.

\bibitem[Fukushima(1980)]{Fukushima_1980}
Kunihiko Fukushima.
\newblock Neocognitron: {A} self-organizing neural network model for a
  mechanism of pattern recognition unaffected by shift in position.
\newblock \emph{Biological Cybernetics}, 36\penalty0 (4):\penalty0 193--202,
  April 1980.
\newblock ISSN 1432-0770.
\newblock \doi{10.1007/BF00344251}.
\newblock URL \url{https://doi.org/10.1007/BF00344251}.

\bibitem[Funke et~al.(2021)Funke, Borowski, Stosio, Brendel, Wallis, and
  Bethge]{Funke-etal_2021}
Christina~M. Funke, Judy Borowski, Karolina Stosio, Wieland Brendel, Thomas
  S.~A. Wallis, and Matthias Bethge.
\newblock Five points to check when comparing visual perception in humans and
  machines.
\newblock \emph{Journal of Vision}, 21\penalty0 (3 (16)):\penalty0 1--23, March
  2021.
\newblock ISSN 1534-7362.
\newblock \doi{10.1167/jov.21.3.16}.
\newblock URL \url{https://doi.org/10.1167/jov.21.3.16}.

\bibitem[Gale et~al.(2020)Gale, Martin, Blything, Nguyen, and
  Bowers]{Gale-etal_2020}
Ella~M. Gale, Nicholas Martin, Ryan Blything, Anh Nguyen, and Jeffrey~S.
  Bowers.
\newblock Are there any ‘object detectors’ in the hidden layers of {CNNs}
  trained to identify objects or scenes?
\newblock \emph{Vision Research}, 176:\penalty0 60--71, November 2020.
\newblock ISSN 0042-6989.
\newblock \doi{10.1016/j.visres.2020.06.007}.
\newblock URL
  \url{https://www.sciencedirect.com/science/article/pii/S0042698920301140}.

\bibitem[Gatys et~al.(2016)Gatys, Ecker, and Bethge]{Gatys2016}
Leon~A Gatys, Alexander~S Ecker, and Matthias Bethge.
\newblock Image style transfer using convolutional neural networks.
\newblock In \emph{{Proceedings of the IEEE Conference on Computer Vision and
  Pattern Recognition}}, pages 2414--2423, 2016.

\bibitem[Geirhos et~al.(2018)Geirhos, Temme, Rauber, Sch{\"u}tt, Bethge, and
  Wichmann]{geirhos2018generalisation}
Robert Geirhos, Carlos~RM Temme, Jonas Rauber, Heiko~H Sch{\"u}tt, Matthias
  Bethge, and Felix~A Wichmann.
\newblock Generalisation in humans and deep neural networks.
\newblock In \emph{{Advances in Neural Information Processing Systems}}, 2018.

\bibitem[Geirhos et~al.(2019)Geirhos, Rubisch, Michaelis, Bethge, Wichmann, and
  Brendel]{geirhos2019imagenettrained}
Robert Geirhos, Patricia Rubisch, Claudio Michaelis, Matthias Bethge, Felix~A.
  Wichmann, and Wieland Brendel.
\newblock {ImageNet}-trained {CNN}s are biased towards texture; increasing
  shape bias improves accuracy and robustness.
\newblock In \emph{{International Conference on Learning Representations}},
  2019.

\bibitem[Geirhos et~al.(2020{\natexlab{a}})Geirhos, Jacobsen, Michaelis, Zemel,
  Brendel, Bethge, and Wichmann]{geirhos2020shortcut}
Robert Geirhos, J{\"o}rn-Henrik Jacobsen, Claudio Michaelis, Richard Zemel,
  Wieland Brendel, Matthias Bethge, and Felix~A Wichmann.
\newblock Shortcut learning in deep neural networks.
\newblock \emph{{Nature Machine Intelligence}}, 2:\penalty0 665–673,
  2020{\natexlab{a}}.

\bibitem[Geirhos et~al.(2020{\natexlab{b}})Geirhos, Meding, and
  Wichmann]{geirhos2020beyond}
Robert Geirhos, Kristof Meding, and Felix~A Wichmann.
\newblock Beyond accuracy: quantifying trial-by-trial behaviour of {CNNs} and
  humans by measuring error consistency.
\newblock \emph{{Advances in Neural Information Processing Systems}}, 33,
  2020{\natexlab{b}}.

\bibitem[Geirhos et~al.(2021)Geirhos, Narayanappa, Mitzkus, Thieringer, Bethge,
  Wichmann, and Brendel]{geirhosPartialSuccessClosing2021}
Robert Geirhos, Kantharaju Narayanappa, Benjamin Mitzkus, Tizian Thieringer,
  Matthias Bethge, Felix~A Wichmann, and Wieland Brendel.
\newblock Partial success in closing the gap between human and machine vision.
\newblock \emph{Advances in Neural Information Processing Systems}, 34, 2021.

\bibitem[Gibson(1950)]{Gibson_1950a}
James~J Gibson.
\newblock \emph{The perception of the visual world}.
\newblock January 1950.

\bibitem[Gigerenzer(1991)]{Gigerenzer_1991}
Gerd Gigerenzer.
\newblock From tools to theories: {{A}} heuristic of discovery in cognitive
  psychology.
\newblock \emph{Psychological Review}, 98\penalty0 (2):\penalty0 254--267,
  1991.
\newblock ISSN 0033-295X.

\bibitem[Goetschalckx et~al.(2021)Goetschalckx, Andonian, and
  Wagemans]{Goetschalckx-etal_2021}
Lore Goetschalckx, Alex Andonian, and Johan Wagemans.
\newblock Generative adversarial networks unlock new methods for cognitive
  science.
\newblock \emph{{Trends in Cognitive Sciences}}, 25\penalty0 (9):\penalty0
  788--801, September 2021.
\newblock ISSN 1364-6613.
\newblock \doi{10.1016/j.tics.2021.06.006}.

\bibitem[Golan et~al.(2020)Golan, Raju, and Kriegeskorte]{Golan-etal_2020a}
Tal Golan, Prashant~C. Raju, and Nikolaus Kriegeskorte.
\newblock Controversial stimuli: {{Pitting}} neural networks against each other
  as models of human cognition.
\newblock \emph{Proceedings of the National Academy of Sciences}, 117\penalty0
  (47):\penalty0 29330--29337, November 2020.
\newblock ISSN 0027-8424, 1091-6490.
\newblock \doi{10.1073/pnas.1912334117}.

\bibitem[Gon{\c c}alves et~al.(2020)Gon{\c c}alves, Lueckmann, Deistler,
  Nonnenmacher, {\"O}cal, Bassetto, Chintaluri, Podlaski, Haddad, Vogels,
  Greenberg, and Macke]{Goncalves-etal_2020}
Pedro~J Gon{\c c}alves, Jan-Matthis Lueckmann, Michael Deistler, Marcel
  Nonnenmacher, Kaan {\"O}cal, Giacomo Bassetto, Chaitanya Chintaluri,
  William~F Podlaski, Sara~A Haddad, Tim~P Vogels, David~S Greenberg, and
  Jakob~H Macke.
\newblock Training deep neural density estimators to identify mechanistic
  models of neural dynamics.
\newblock \emph{eLife}, 9:\penalty0 e56261, September 2020.
\newblock ISSN 2050-084X.
\newblock \doi{10.7554/eLife.56261}.

\bibitem[Goodfellow et~al.(2016)Goodfellow, Bengio, and
  Courville]{goodfellow2016deep}
Ian Goodfellow, Yoshua Bengio, and Aaron Courville.
\newblock \emph{Deep learning}.
\newblock MIT press, 2016.

\bibitem[Goris et~al.(2013)Goris, Putzeys, Wagemans, and Wichmann]{Goris2013}
Robbe~LT Goris, Tom Putzeys, Johan Wagemans, and Felix~A Wichmann.
\newblock A neural population model for visual pattern detection.
\newblock \emph{Psychological Review}, 120\penalty0 (3):\penalty0 472, 2013.

\bibitem[Green(1964)]{green1964consistency}
David~M. Green.
\newblock Consistency of auditory detection judgments.
\newblock \emph{Psychological Review}, 71\penalty0 (5):\penalty0 392--407,
  1964.

\bibitem[Hassenstein and Reichardt(1956)]{Hassenstein_1956}
B~Hassenstein and W~Reichardt.
\newblock Systemtheoretische {Analyse} der {Zeit}, {Reihenfolgen} und
  {Vorzeichenauswertung} bei der {Bewegungsrezeption} des {R}üsselkäfers
  {Chlorophanus}.
\newblock \emph{Zeitschrift für Naturforschung, Teil B}, 11\penalty0
  (9-10):\penalty0 513, January 1956.

\bibitem[{He} et~al.(2015){He}, Zhang, Ren, and Sun]{he2015delving}
Kaiming {He}, Xiangyu Zhang, Shaoqing Ren, and Jian Sun.
\newblock Delving deep into rectifiers: Surpassing human-level performance on
  {ImageNet} classification.
\newblock In \emph{{Proceedings of the IEEE International Conference on
  Computer Vision}}, pages 1026--1034, 2015.

\bibitem[Hein et~al.(2019)Hein, Andriushchenko, and
  Bitterwolf]{Hein-etal_2019a}
Matthias Hein, Maksym Andriushchenko, and Julian Bitterwolf.
\newblock Why {{ReLU Networks Yield High-Confidence Predictions Far Away From}}
  the {{Training Data}} and {{How}} to {{Mitigate}} the {{Problem}}.
\newblock In \emph{2019 {{IEEE}}/{{CVF Conference}} on {{Computer Vision}} and
  {{Pattern Recognition}} ({{CVPR}})}, pages 41--50, June 2019.
\newblock \doi{10.1109/CVPR.2019.00013}.

\bibitem[Hendrycks and Dietterich(2019)]{hendrycks2019benchmarking}
Dan Hendrycks and Thomas Dietterich.
\newblock Benchmarking neural network robustness to common corruptions and
  perturbations.
\newblock In \emph{{International Conference on Learning Representations}},
  2019.

\bibitem[Hendrycks et~al.(2021{\natexlab{a}})Hendrycks, Basart, Mu, Kadavath,
  Wang, Dorundo, Desai, Zhu, Parajuli, Guo, et~al.]{hendrycks2021many}
Dan Hendrycks, Steven Basart, Norman Mu, Saurav Kadavath, Frank Wang, Evan
  Dorundo, Rahul Desai, Tyler Zhu, Samyak Parajuli, Mike Guo, et~al.
\newblock The many faces of robustness: A critical analysis of
  out-of-distribution generalization.
\newblock In \emph{{Proceedings of the IEEE Conference on Computer Vision and
  Pattern Recognition}}, pages 8340--8349, 2021{\natexlab{a}}.

\bibitem[Hendrycks et~al.(2021{\natexlab{b}})Hendrycks, Zhao, Basart,
  Steinhardt, and Song]{hendrycks2021natural}
Dan Hendrycks, Kevin Zhao, Steven Basart, Jacob Steinhardt, and Dawn Song.
\newblock Natural adversarial examples.
\newblock In \emph{{Proceedings of the IEEE Conference on Computer Vision and
  Pattern Recognition}}, pages 15262--15271, 2021{\natexlab{b}}.

\bibitem[Hermann et~al.(2020)Hermann, Chen, and Kornblith]{hermann2020the}
Katherine Hermann, Ting Chen, and Simon Kornblith.
\newblock The origins and prevalence of texture bias in convolutional neural
  networks.
\newblock \emph{{Advances in Neural Information Processing Systems}}, 33, 2020.

\bibitem[Hornik(1991)]{Hornik_1991}
Kurt Hornik.
\newblock Approximation capabilities of multilayer feedforward networks.
\newblock \emph{Neural Networks}, 4\penalty0 (2):\penalty0 251--257, 1991.
\newblock ISSN 0893-6080.

\bibitem[Hornik et~al.(1989)Hornik, Stinchcombe, and
  White]{hornik1989multilayer}
Kurt Hornik, Maxwell Stinchcombe, and Halbert White.
\newblock Multilayer feedforward networks are universal approximators.
\newblock \emph{{Neural Networks}}, 2\penalty0 (5):\penalty0 359--366, 1989.

\bibitem[Huang and Belongie(2017)]{Huang-Belongie_2017}
Xun Huang and Serge Belongie.
\newblock Arbitrary style transfer in real-time with adaptive instance
  normalization.
\newblock In \emph{2017 {IEEE} {International} {Conference} on {Computer}
  {Vision} ({ICCV})}, pages 1510--1519, October 2017.
\newblock \doi{10.1109/ICCV.2017.167}.
\newblock ISSN: 2380-7504.

\bibitem[Huber et~al.(2022)Huber, Geirhos, and
  Wichmann]{huber2022developmental}
Lukas~S Huber, Robert Geirhos, and Felix~A Wichmann.
\newblock The developmental trajectory of object recognition robustness:
  children are like small adults but unlike big deep neural networks.
\newblock \emph{arXiv:2205.10144}, 2022.

\bibitem[Ibrahim et~al.(2022)Ibrahim, Garrido, Morcos, and
  Bouchacourt]{ibrahim2022robustness}
Mark Ibrahim, Quentin Garrido, Ari Morcos, and Diane Bouchacourt.
\newblock The robustness limits of sota vision models to natural variation.
\newblock \emph{arXiv:2210.13604}, 2022.

\bibitem[Idrissi et~al.(2022)Idrissi, Bouchacourt, Balestriero, Evtimov,
  Hazirbas, Ballas, Vincent, Drozdzal, Lopez-Paz, and
  Ibrahim]{Idrissi-etal_2022}
Badr~Youbi Idrissi, Diane Bouchacourt, Randall Balestriero, Ivan Evtimov, Caner
  Hazirbas, Nicolas Ballas, Pascal Vincent, Michal Drozdzal, David Lopez-Paz,
  and Mark Ibrahim.
\newblock {ImageNet}-{X}: {Understanding} {Model} {Mistakes} with {Factor} of
  {Variation} {Annotations}, November 2022.
\newblock URL \url{http://arxiv.org/abs/2211.01866}.
\newblock arXiv:2211.01866 [cs].

\bibitem[Jastrow(1899)]{jastrow1899mind}
Joseph Jastrow.
\newblock The mind's eye.
\newblock \emph{Popular Science Monthly}, 1899.

\bibitem[Kietzmann et~al.(2019)Kietzmann, Spoerer, S{\"o}rensen, Cichy, Hauk,
  and Kriegeskorte]{kietzmann2019recurrence}
Tim~C Kietzmann, Courtney~J Spoerer, Lynn~KA S{\"o}rensen, Radoslaw~M Cichy,
  Olaf Hauk, and Nikolaus Kriegeskorte.
\newblock Recurrence is required to capture the representational dynamics of
  the human visual system.
\newblock \emph{Proceedings of the National Academy of Sciences}, 116\penalty0
  (43):\penalty0 21854--21863, 2019.

\bibitem[Kindermans et~al.(2017)Kindermans, Hooker, Adebayo, Alber, Sch{\"u}tt,
  D{\"a}hne, Erhan, and Kim]{Kindermans-etal_2017}
Pieter-Jan Kindermans, Sara Hooker, Julius Adebayo, Maximilian Alber,
  Kristof~T. Sch{\"u}tt, Sven D{\"a}hne, Dumitru Erhan, and Been Kim.
\newblock The ({{Un}})reliability of saliency methods, November 2017.

\bibitem[Koenderink et~al.(2017)Koenderink, Valsecchi, Doorn, Wagemans, and
  Gegenfurtner]{Koenderink_2017}
Jan Koenderink, Matteo Valsecchi, Andrea~van Doorn, Johan Wagemans, and Karl
  Gegenfurtner.
\newblock Eidolons: {Novel} stimuli for vision research.
\newblock \emph{Journal of Vision}, 17\penalty0 (2):\penalty0 7, 1--36, 2017.
\newblock ISSN 1534-7362.
\newblock URL \url{https://doi.org/10.1167/17.2.7}.

\bibitem[Koh et~al.(2021)Koh, Sagawa, Marklund, Xie, Zhang, Balsubramani, Hu,
  Yasunaga, Phillips, Gao, et~al.]{koh2021wilds}
Pang~Wei Koh, Shiori Sagawa, Henrik Marklund, Sang~Michael Xie, Marvin Zhang,
  Akshay Balsubramani, Weihua Hu, Michihiro Yasunaga, Richard~Lanas Phillips,
  Irena Gao, et~al.
\newblock Wilds: A benchmark of in-the-wild distribution shifts.
\newblock In \emph{International Conference on Machine Learning}, pages
  5637--5664. PMLR, 2021.

\bibitem[Kosiorek et~al.(2019)Kosiorek, Sabour, Teh, and
  Hinton]{kosiorek2019stacked}
Adam~R Kosiorek, Sara Sabour, Yee~Whye Teh, and Geoffrey~E Hinton.
\newblock Stacked capsule autoencoders.
\newblock \emph{arXiv:1906.06818}, 2019.

\bibitem[Kreiman and Serre(2020)]{kreiman2020beyond}
Gabriel Kreiman and Thomas Serre.
\newblock Beyond the feedforward sweep: feedback computations in the visual
  cortex.
\newblock \emph{Annals of the New York Academy of Sciences}, 1464\penalty0
  (1):\penalty0 222--241, 2020.

\bibitem[Kriegeskorte(2015)]{kriegeskorte2015deep}
Nikolaus Kriegeskorte.
\newblock Deep neural networks: a new framework for modeling biological vision
  and brain information processing.
\newblock \emph{{Annual Review of Vision Science}}, 1:\penalty0 417--446, 2015.

\bibitem[Krizhevsky et~al.(2012)Krizhevsky, Sutskever, and
  Hinton]{krizhevsky2012imagenet}
Alex Krizhevsky, Ilya Sutskever, and Geoffrey~E Hinton.
\newblock {ImageNet} classification with deep convolutional neural networks.
\newblock In \emph{{Advances in Neural Information Processing Systems}}, pages
  1097--1105, 2012.

\bibitem[Kubilius et~al.(2019)Kubilius, Schrimpf, Kar, Rajalingham, Hong,
  Majaj, Issa, Bashivan, Prescott-Roy, Schmidt, et~al.]{kubilius2019brain}
Jonas Kubilius, Martin Schrimpf, Kohitij Kar, Rishi Rajalingham, Ha~Hong, Najib
  Majaj, Elias Issa, Pouya Bashivan, Jonathan Prescott-Roy, Kailyn Schmidt,
  et~al.
\newblock Brain-like object recognition with high-performing shallow recurrent
  {ANNs}.
\newblock \emph{{Advances in Neural Information Processing Systems}},
  32:\penalty0 12805--12816, 2019.

\bibitem[Lake et~al.(2017)Lake, Ullman, Tenenbaum, and Gershman]{Lake_2017}
Brenden~M. Lake, Tomer~D. Ullman, Joshua~B. Tenenbaum, and Samuel~J. Gershman.
\newblock Building machines that learn and think like people.
\newblock \emph{Behavioral and Brain Sciences}, 40:\penalty0 e253, 2017.
\newblock ISSN 0140-525X.

\bibitem[Lauer et~al.(2022)Lauer, Zhou, Ye, Menegas, Schneider, Nath, Rahman,
  Di~Santo, Soberanes, Feng, Murthy, Lauder, Dulac, Mathis, and
  Mathis]{Lauer-etal_2022}
Jessy Lauer, Mu~Zhou, Shaokai Ye, William Menegas, Steffen Schneider, Tanmay
  Nath, Mohammed~Mostafizur Rahman, Valentina Di~Santo, Daniel Soberanes,
  Guoping Feng, Venkatesh~N. Murthy, George Lauder, Catherine Dulac,
  Mackenzie~Weygandt Mathis, and Alexander Mathis.
\newblock Multi-animal pose estimation, identification and tracking with
  {{DeepLabCut}}.
\newblock \emph{Nature Methods}, 19\penalty0 (4):\penalty0 496--504, April
  2022.
\newblock ISSN 1548-7105.
\newblock \doi{10.1038/s41592-022-01443-0}.

\bibitem[LeCun et~al.(1989)LeCun, Boser, Denker, Henderson, Howard, Hubbard,
  and Jackel]{LeCun-etal_1989}
Y.~LeCun, B.~Boser, J.~S. Denker, D.~Henderson, R.~E. Howard, W.~Hubbard, and
  L.~D. Jackel.
\newblock Backpropagation {Applied} to {Handwritten} {Zip} {Code}
  {Recognition}.
\newblock \emph{Neural Computation}, 1\penalty0 (4):\penalty0 541--551,
  December 1989.
\newblock ISSN 0899-7667.
\newblock \doi{10.1162/neco.1989.1.4.541}.
\newblock URL \url{https://doi.org/10.1162/neco.1989.1.4.541}.

\bibitem[LeCun et~al.(2015)LeCun, Bengio, and Hinton]{lecun2015deep}
Yann LeCun, Yoshua Bengio, and Geoffrey Hinton.
\newblock Deep learning.
\newblock \emph{{Nature}}, 521\penalty0 (7553):\penalty0 436, 2015.

\bibitem[Logothetis and Sheinberg(1996)]{Logothetis_1996}
N~K Logothetis and D~L Sheinberg.
\newblock Visual object recognition.
\newblock \emph{Annual Review of Neuroscience}, 19\penalty0 (1):\penalty0
  577--621, 1996.
\newblock ISSN 0147-006x.

\bibitem[Lonnqvist et~al.(2021)Lonnqvist, Bornet, Doerig, and
  Herzog]{Lonnqvist-etal_2021a}
Ben Lonnqvist, Alban Bornet, Adrien Doerig, and Michael~H. Herzog.
\newblock A comparative biology approach to {{DNN}} modeling of vision: {{A}}
  focus on differences, not similarities.
\newblock \emph{Journal of Vision}, 21\penalty0 (10):\penalty0 17--17,
  September 2021.
\newblock ISSN 1534-7362.
\newblock \doi{10.1167/jov.21.10.17}.

\bibitem[Ma and Peters(2020)]{ma2020neural}
Wei~Ji Ma and Benjamin Peters.
\newblock A neural network walks into a lab: towards using deep nets as models
  for human behavior.
\newblock \emph{arXiv:2005.02181}, 2020.

\bibitem[Madry et~al.(2017)Madry, Makelov, Schmidt, Tsipras, and
  Vladu]{madry2017towards}
Aleksander Madry, Aleksandar Makelov, Ludwig Schmidt, Dimitris Tsipras, and
  Adrian Vladu.
\newblock Towards deep learning models resistant to adversarial attacks.
\newblock \emph{arXiv:1706.06083}, 2017.

\bibitem[Malhotra et~al.(2022)Malhotra, Dujmović, and
  Bowers]{Malhotra-etal_2022}
Gaurav Malhotra, Marin Dujmović, and Jeffrey~S. Bowers.
\newblock Feature blindness: {A} challenge for understanding and modelling
  visual object recognition.
\newblock \emph{{PLoS Computational Biology}}, 18\penalty0 (5):\penalty0
  e1009572, May 2022.
\newblock ISSN 1553-7358.
\newblock \doi{10.1371/journal.pcbi.1009572}.
\newblock URL
  \url{https://journals.plos.org/ploscompbiol/article?id=10.1371/journal.pcbi.1009572}.

\bibitem[Marcel and Rodriguez(2010)]{marcel2010torchvision}
S{\'e}bastien Marcel and Yann Rodriguez.
\newblock Torchvision the machine-vision package of torch.
\newblock In \emph{{Proceedings of the 18th ACM International Conference on
  Multimedia}}, pages 1485--1488, 2010.

\bibitem[Marcus(2018)]{marcus2018deep}
Gary Marcus.
\newblock Deep learning: A critical appraisal.
\newblock \emph{arXiv:1801.00631}, 2018.

\bibitem[Mathis et~al.(2018)Mathis, Mamidanna, Cury, Abe, Murthy, Mathis, and
  Bethge]{Mathis-etal_2018}
Alexander Mathis, Pranav Mamidanna, Kevin~M. Cury, Taiga Abe, Venkatesh~N.
  Murthy, Mackenzie~Weygandt Mathis, and Matthias Bethge.
\newblock {{DeepLabCut}}: markerless pose estimation of user-defined body parts
  with deep learning.
\newblock \emph{Nature Neuroscience}, 21\penalty0 (9):\penalty0 1281--1289,
  September 2018.
\newblock ISSN 1546-1726.
\newblock \doi{10.1038/s41593-018-0209-y}.

\bibitem[McClelland et~al.(1986)McClelland, Rumelhart, and
  Group]{McClelland-etal_1986}
James~L. McClelland, David~E. Rumelhart, and PDP~Research Group, editors.
\newblock \emph{Parallel {{Distributed Processing}}, {{Volume II}}:
  {{Explorations}} in the {{Microstructure}} of {{Cognition}}:
  {{Psychological}} and {{Biological Models}}}.
\newblock {MIT Press}, {Cambridge, MA}, 1986.

\bibitem[McCulloch and Pitts(1943)]{McCulloch-Pitts_1943}
Warren~S. McCulloch and Walter Pitts.
\newblock A logical calculus of the ideas immanent in nervous activity.
\newblock \emph{The bulletin of mathematical biophysics}, 5\penalty0
  (4):\penalty0 115--133, December 1943.
\newblock ISSN 1522-9602.
\newblock \doi{10.1007/BF02478259}.

\bibitem[Meding et~al.(2022)Meding, Buschoff, Geirhos, and
  Wichmann]{meding2021trivial}
Kristof Meding, Luca M~Schulze Buschoff, Robert Geirhos, and Felix~A Wichmann.
\newblock Trivial or impossible--dichotomous data difficulty masks model
  differences (on {ImageNet} and beyond).
\newblock \emph{International Conference on Learning Representations (ICLR)},
  2022.

\bibitem[Mehrer et~al.(2021)Mehrer, Spoerer, Jones, Kriegeskorte, and
  Kietzmann]{Mehrer-etal_2021}
Johannes Mehrer, Courtney~J. Spoerer, Emer~C. Jones, Nikolaus Kriegeskorte, and
  Tim~C. Kietzmann.
\newblock An ecologically motivated image dataset for deep learning yields
  better models of human vision.
\newblock \emph{Proceedings of the National Academy of Sciences}, 118\penalty0
  (8):\penalty0 e2011417118, February 2021.
\newblock \doi{10.1073/pnas.2011417118}.
\newblock URL \url{https://www.pnas.org/doi/10.1073/pnas.2011417118}.

\bibitem[Mitchell(1980)]{Mitchell_1980}
Tom~M. Mitchell.
\newblock The need for biases in learning generalizations.
\newblock \emph{Rutgers CS Tech Report}, CBM-TR-117:\penalty0 1--3, 1980.

\bibitem[Montavon et~al.(2017)Montavon, Lapuschkin, Binder, Samek, and
  M{\"u}ller]{Montavon-etal_2017}
Gr{\'e}goire Montavon, Sebastian Lapuschkin, Alexander Binder, Wojciech Samek,
  and Klaus-Robert M{\"u}ller.
\newblock Explaining nonlinear classification decisions with deep {{Taylor}}
  decomposition.
\newblock \emph{Pattern Recognition}, 65:\penalty0 211--222, May 2017.
\newblock ISSN 0031-3203.
\newblock \doi{10.1016/j.patcog.2016.11.008}.

\bibitem[Nakayama et~al.(1995)Nakayama, He, and Shimojo]{Nakayama_1995}
Ken Nakayama, Zijiang~J. He, and Shinsuke Shimojo.
\newblock Visual surface representation: {{A}} critical link between
  lower-level and higher-level vision.
\newblock In Stephen~M Kosslyn and D.~N. Osherson, editors, \emph{An
  {{Invitation}} to {{Cognitive Science}}}, pages 1--70. {MIT Press}, 1995.

\bibitem[Oliva et~al.(2006)Oliva, Torralba, and Schyns]{oliva2006hybrid}
Aude Oliva, Antonio Torralba, and Philippe~G Schyns.
\newblock Hybrid images.
\newblock \emph{ACM Transactions on Graphics (TOG)}, 25\penalty0 (3):\penalty0
  527--532, 2006.

\bibitem[O'Toole and Castillo(2021)]{OToole-Castillo_2021}
Alice~J. O'Toole and Carlos~D. Castillo.
\newblock Face {Recognition} by {Humans} and {Machines}: {Three} {Fundamental}
  {Advances} from {Deep} {Learning}.
\newblock \emph{Annual Review of Vision Science}, 7\penalty0 (1):\penalty0
  543--570, September 2021.
\newblock ISSN 2374-4642, 2374-4650.
\newblock \doi{10.1146/annurev-vision-093019-111701}.
\newblock URL
  \url{https://www.annualreviews.org/doi/10.1146/annurev-vision-093019-111701}.

\bibitem[Peissig and Tarr(2007)]{Peissig_2007}
Jessie~J. Peissig and Michael~J. Tarr.
\newblock Visual object recognition: {D}o we know more now than we did 20 years
  ago?
\newblock \emph{Annual Review of Psychology}, 58\penalty0 (1):\penalty0 75--96,
  2007.
\newblock ISSN 0066-4308.
\newblock \doi{10.1146/annurev.psych.58.102904.190114}.

\bibitem[Peters and Kriegeskorte(2021)]{Peters-Kriegeskorte_2021b}
Benjamin Peters and Nikolaus Kriegeskorte.
\newblock Capturing the objects of vision with neural networks.
\newblock \emph{Nature Human Behaviour}, 5\penalty0 (9):\penalty0 1127--1144,
  September 2021.
\newblock ISSN 2397-3374.
\newblock \doi{10.1038/s41562-021-01194-6}.

\bibitem[Piloto et~al.(2022)Piloto, Weinstein, Battaglia, and
  Botvinick]{Piloto-etal_2022}
Luis~S. Piloto, Ari Weinstein, Peter Battaglia, and Matthew Botvinick.
\newblock Intuitive physics learning in a deep-learning model inspired by
  developmental psychology.
\newblock \emph{Nature Human Behaviour}, 6\penalty0 (9):\penalty0 1257--1267,
  September 2022.
\newblock ISSN 2397-3374.
\newblock \doi{10.1038/s41562-022-01394-8}.

\bibitem[Rajalingham et~al.(2018)Rajalingham, Issa, Bashivan, Kar, Schmidt, and
  DiCarlo]{rajalingham2018large}
Rishi Rajalingham, Elias~B Issa, Pouya Bashivan, Kohitij Kar, Kailyn Schmidt,
  and James~J DiCarlo.
\newblock Large-scale, high-resolution comparison of the core visual object
  recognition behavior of humans, monkeys, and state-of-the-art deep artificial
  neural networks.
\newblock \emph{Journal of Neuroscience}, 38\penalty0 (33):\penalty0
  7255--7269, 2018.

\bibitem[Ramadhan et~al.(2022)Ramadhan, Marshall, Souza, Lee, Piterbarg,
  Hillier, Wagner, Rackauckas, Hill, Campin, and Ferrari]{Ramadhan-etal_2022}
Ali Ramadhan, John~C. Marshall, Andre~Nogueira Souza, Xin~Kai Lee, Ulyana
  Piterbarg, Adeline Hillier, Gregory~LeClaire Wagner, Christopher Rackauckas,
  Chris Hill, Jean-Michel Campin, and Raffaele Ferrari.
\newblock Capturing missing physics in climate model parameterizations using
  neural differential equations, October 2022.

\bibitem[Rauber et~al.(2017)Rauber, Brendel, and Bethge]{rauber2017foolbox}
Jonas Rauber, Wieland Brendel, and Matthias Bethge.
\newblock Foolbox: A python toolbox to benchmark the robustness of machine
  learning models.
\newblock In \emph{Reliable Machine Learning in the Wild Workshop, 34th
  International Conference on Machine Learning}, 2017.
\newblock URL \url{http://arxiv.org/abs/1707.04131}.

\bibitem[Recht et~al.(2019)Recht, Roelofs, Schmidt, and
  Shankar]{Recht-etal_2019}
Benjamin Recht, Rebecca Roelofs, Ludwig Schmidt, and Vaishaal Shankar.
\newblock Do {{ImageNet classifiers generalize}} to {{ImageNet}}?
\newblock In \emph{Proceedings of the 36th {{International Conference}} on
  {{Machine Learning}}}, pages 5389--5400. {PMLR}, May 2019.

\bibitem[Reichardt(1957)]{Reichardt_1957}
W~Reichardt.
\newblock Autokorrelationsauswertung als {Funktionsprinzip} des
  {Zentralnervensystems}.
\newblock \emph{Zeitschrift für Naturforschung, Teil B}, 12:\penalty0 447,
  January 1957.

\bibitem[Rideaux et~al.(2021)Rideaux, Storrs, Maiello, and
  Welchman]{Rideaux-etal_2021}
Reuben Rideaux, Katherine~R. Storrs, Guido Maiello, and Andrew~E. Welchman.
\newblock How multisensory neurons solve causal inference.
\newblock \emph{Proceedings of the National Academy of Sciences}, 118\penalty0
  (32):\penalty0 e2106235118, August 2021.
\newblock \doi{10.1073/pnas.2106235118}.

\bibitem[Riesenhuber and Poggio(1999)]{riesenhuberHierarchicalModelsObject1999}
Maximilian Riesenhuber and Tomaso Poggio.
\newblock Hierarchical models of object recognition in cortex.
\newblock \emph{Nature Neuroscience}, 2\penalty0 (11):\penalty0 1019--1025,
  1999.
\newblock ISSN 1097-6256.
\newblock URL \url{https://doi.org/10.1038/14819}.

\bibitem[Rosenblatt(1958)]{Rosenblatt_1958}
F.~Rosenblatt.
\newblock The perceptron: a probabilistic model for information storage and
  organization in the brain.
\newblock \emph{Psychological Review}, 65\penalty0 (6):\penalty0 386--408,
  November 1958.
\newblock ISSN 0033-295X.
\newblock \doi{10.1037/h0042519}.

\bibitem[Rumelhart et~al.(1986)Rumelhart, McClelland, and
  Group]{Rumelhart-etal_1986}
David~E. Rumelhart, James~L. McClelland, and PDP~Research Group, editors.
\newblock \emph{Parallel {{Distributed Processing}}, {{Volume}} 1:
  {{Explorations}} in the {{Microstructure}} of {{Cognition}}:
  {{Foundations}}}.
\newblock {MIT Press}, {Cambridge, MA}, 1986.

\bibitem[Sabour et~al.(2017)Sabour, Frosst, and Hinton]{sabour2017dynamic}
Sara Sabour, Nicholas Frosst, and Geoffrey~E Hinton.
\newblock Dynamic routing between capsules.
\newblock \emph{Advances in Neural Information Processing Systems}, 30, 2017.

\bibitem[Safavi and Dayan(2022)]{Safavi-Dayan_2022}
Shervin Safavi and Peter Dayan.
\newblock Multistability, perceptual value, and internal foraging.
\newblock \emph{Neuron}, 110\penalty0 (19):\penalty0 3076--3090, October 2022.
\newblock ISSN 0896-6273.
\newblock \doi{10.1016/j.neuron.2022.07.024}.

\bibitem[Schade(1956)]{Schade_1956}
O~H Schade.
\newblock Optical and photoelectric analogue of the eye.
\newblock \emph{Journal of the Optical Society of America}, 46:\penalty0
  721--739, January 1956.

\bibitem[Schmidhuber(2015)]{Schmidhuber_2015}
J{\"u}rgen Schmidhuber.
\newblock Deep learning in neural networks: {{An}} overview.
\newblock \emph{Neural Networks}, 61:\penalty0 85--117, 2015.
\newblock ISSN 0893-6080.

\bibitem[Schütt and Wichmann(2017)]{Schutt_2017}
Heiko~H. Schütt and Felix~A. Wichmann.
\newblock An image-computable psychophysical spatial vision model.
\newblock \emph{Journal of Vision}, 17\penalty0 (12):\penalty0 12, 1--35, 2017.
\newblock ISSN 1534-7362.
\newblock URL \url{https://doi.org/10.1167/17.12.12}.

\bibitem[Senior et~al.(2020)Senior, Evans, Jumper, Kirkpatrick, Sifre, Green,
  Qin, {\v Z}{\'i}dek, Nelson, Bridgland, Penedones, Petersen, Simonyan,
  Crossan, Kohli, Jones, Silver, Kavukcuoglu, and Hassabis]{Senior-etal_2020}
Andrew~W. Senior, Richard Evans, John Jumper, James Kirkpatrick, Laurent Sifre,
  Tim Green, Chongli Qin, Augustin {\v Z}{\'i}dek, Alexander W.~R. Nelson, Alex
  Bridgland, Hugo Penedones, Stig Petersen, Karen Simonyan, Steve Crossan,
  Pushmeet Kohli, David~T. Jones, David Silver, Koray Kavukcuoglu, and Demis
  Hassabis.
\newblock Improved protein structure prediction using potentials from deep
  learning.
\newblock \emph{{Nature}}, 577\penalty0 (7792):\penalty0 706--710, January
  2020.
\newblock ISSN 1476-4687.
\newblock \doi{10.1038/s41586-019-1923-7}.

\bibitem[Serre(2019)]{serre2019deep}
Thomas Serre.
\newblock Deep learning: the good, the bad, and the ugly.
\newblock \emph{{Annual Review of Vision Science}}, 5:\penalty0 399--426, 2019.

\bibitem[Singh et~al.(2022)Singh, Gustafson, Adcock, Reis, Gedik, Kosaraju,
  Mahajan, Girshick, Doll{\'a}r, and van~der Maaten]{singh2022revisiting}
Mannat Singh, Laura Gustafson, Aaron Adcock, Vinicius de~Freitas Reis, Bugra
  Gedik, Raj~Prateek Kosaraju, Dhruv Mahajan, Ross Girshick, Piotr Doll{\'a}r,
  and Laurens van~der Maaten.
\newblock Revisiting weakly supervised pre-training of visual perception
  models.
\newblock \emph{arXiv:2201.08371}, 2022.

\bibitem[Speiser et~al.(2021)Speiser, M{\"u}ller, Hoess, Matti, Obara, Legant,
  Kreshuk, Macke, Ries, and Turaga]{Speiser-etal_2021}
Artur Speiser, Lucas-Raphael M{\"u}ller, Philipp Hoess, Ulf Matti,
  Christopher~J. Obara, Wesley~R. Legant, Anna Kreshuk, Jakob~H. Macke, Jonas
  Ries, and Srinivas~C. Turaga.
\newblock Deep learning enables fast and dense single-molecule localization
  with high accuracy.
\newblock \emph{Nature Methods}, 18\penalty0 (9):\penalty0 1082--1090,
  September 2021.
\newblock ISSN 1548-7105.
\newblock \doi{10.1038/s41592-021-01236-x}.

\bibitem[Storrs and Fleming(2021)]{Storrs_2021b}
Katherine~R. Storrs and Roland~W. Fleming.
\newblock Learning about the world by learning about images.
\newblock \emph{Current Directions in Psychological Science}, 30\penalty0
  (2):\penalty0 120--128, 2021.

\bibitem[Storrs et~al.(2021)Storrs, Anderson, and
  Fleming]{storrs2021unsupervised}
Katherine~R Storrs, Barton~L Anderson, and Roland~W Fleming.
\newblock Unsupervised learning predicts human perception and misperception of
  gloss.
\newblock \emph{{Nature Human Behaviour}}, pages 1--16, 2021.

\bibitem[Szegedy et~al.(2013)Szegedy, Zaremba, Sutskever, Bruna, Erhan,
  Goodfellow, and Fergus]{szegedy2013intriguing}
Christian Szegedy, Wojciech Zaremba, Ilya Sutskever, Joan Bruna, Dumitru Erhan,
  Ian Goodfellow, and Rob Fergus.
\newblock Intriguing properties of neural networks.
\newblock \emph{arXiv:1312.6199}, 2013.

\bibitem[Teller(1984)]{Teller_1984}
Davida~Y. Teller.
\newblock Linking propositions.
\newblock \emph{Vision Research}, 24\penalty0 (10):\penalty0 1233--1246,
  January 1984.
\newblock ISSN 0042-6989.
\newblock \doi{10.1016/0042-6989(84)90178-0}.
\newblock URL
  \url{https://www.sciencedirect.com/science/article/pii/0042698984901780}.

\bibitem[Todorovi{\'c}(2020)]{todorovic2020visual}
Dejan Todorovi{\'c}.
\newblock What are visual illusions?
\newblock \emph{Perception}, 49\penalty0 (11):\penalty0 1128--1199, 2020.

\bibitem[Torralba and Efros(2011)]{torralba2011unbiased}
Antonio Torralba and Alexei~A Efros.
\newblock Unbiased look at dataset bias.
\newblock In \emph{{Proceedings of the IEEE Conference on Computer Vision and
  Pattern Recognition}}, 2011.

\bibitem[{van Bergen} and Kriegeskorte(2020)]{vanBergen-Kriegeskorte_2020}
Ruben~S {van Bergen} and Nikolaus Kriegeskorte.
\newblock Going in circles is the way forward: The role of recurrence in visual
  inference.
\newblock \emph{Current Opinion in Neurobiology}, 65:\penalty0 176--193,
  December 2020.
\newblock ISSN 0959-4388.
\newblock \doi{10.1016/j.conb.2020.11.009}.

\bibitem[Wang and Simoncelli(2008)]{Wang_2008}
Zhou Wang and Eero~P. Simoncelli.
\newblock Maximum differentiation ({MAD}) competition: {A} methodology for
  comparing computational models of perceptual quantities.
\newblock \emph{Journal of Vision}, 8\penalty0 (12):\penalty0 8--8, 2008.
\newblock ISSN 1534-7362.
\newblock URL \url{https://doi.org/10.1167/8.12.8}.

\bibitem[Wichmann et~al.(2010)Wichmann, Drewes, Rosas, and
  Gegenfurtner]{wichmann2010animal}
Felix~A Wichmann, Jan Drewes, Pedro Rosas, and Karl~R Gegenfurtner.
\newblock Animal detection in natural scenes: Critical features revisited.
\newblock \emph{Journal of Vision}, 10\penalty0 (4):\penalty0 6--6, 2010.

\bibitem[Wichmann et~al.(2017)Wichmann, Janssen, Geirhos, Aguilar, Sch{\"u}tt,
  Maertens, and Bethge]{wichmann2017methods}
Felix~A Wichmann, David~HJ Janssen, Robert Geirhos, Guillermo Aguilar, Heiko~H
  Sch{\"u}tt, Marianne Maertens, and Matthias Bethge.
\newblock Methods and measurements to compare men against machines.
\newblock \emph{Electronic Imaging, Human Vision and Electronic Imaging},
  2017\penalty0 (14):\penalty0 36--45, 2017.

\bibitem[Wolpert and Macready(1997)]{wolpert1997no}
David~H Wolpert and William~G Macready.
\newblock No free lunch theorems for optimization.
\newblock \emph{{IEEE Transactions on Evolutionary Computation}}, 1\penalty0
  (1):\penalty0 67--82, 1997.

\bibitem[Yamins and DiCarlo(2016)]{Yamins2016a}
Daniel~LK Yamins and James~J DiCarlo.
\newblock Using goal-driven deep learning models to understand sensory cortex.
\newblock \emph{Nature Neuroscience}, 19\penalty0 (3):\penalty0 356--365, 2016.

\bibitem[Yamins et~al.(2014)Yamins, Hong, Cadieu, Solomon, Seibert, and
  DiCarlo]{Yamins2014}
Daniel~LK Yamins, Ha~Hong, Charles~F Cadieu, Ethan~A Solomon, Darren Seibert,
  and James~J DiCarlo.
\newblock Performance-optimized hierarchical models predict neural responses in
  higher visual cortex.
\newblock \emph{{Proceedings of the National Academy of Sciences}},
  111\penalty0 (23):\penalty0 8619--8624, 2014.

\bibitem[Zador(2019)]{Zador_2019}
Anthony~M. Zador.
\newblock A critique of pure learning and what artificial neural networks can
  learn from animal brains.
\newblock \emph{{Nature Communications}}, 10\penalty0 (1):\penalty0 3770, 2019.

\bibitem[Zednik and Jäkel(2016)]{Zednik_2016}
Carlos Zednik and Frank Jäkel.
\newblock Bayesian reverse-engineering considered as a research strategy for
  cognitive science.
\newblock \emph{Synthese}, 193\penalty0 (12):\penalty0 3951--3985, 2016.
\newblock URL \url{https://doi.org/10.1007/s11229-016-1180-3}.

\bibitem[Zhou and Firestone(2019)]{zhou2019humans}
Zhenglong Zhou and Chaz Firestone.
\newblock Humans can decipher adversarial images.
\newblock \emph{{Nature Communications}}, 10\penalty0 (1):\penalty0 1334, 2019.

\bibitem[Zimmermann et~al.(2021)Zimmermann, Borowski, Geirhos, Bethge, Wallis,
  and Brendel]{Zimmermann-etal_2021}
Roland~S Zimmermann, Judy Borowski, Robert Geirhos, Matthias Bethge, Thomas S~A
  Wallis, and Wieland Brendel.
\newblock How {Well} do {Feature} {Visualizations} {Support} {Causal}
  {Understanding} of {CNN} {Activations}?
\newblock \emph{Advances in Neural Information Processing Systems},
  34:\penalty0 11730--11744, 2021.

\end{thebibliography}
	\end{small}
	
\end{document}